\pdfoutput=1
\documentclass[11pt]{article}

\usepackage[final]{acl} % Using 'final' as per your example

\usepackage{times}        % Provides Times Roman font
\usepackage{latexsym}     % Provides some LaTeX symbols

\usepackage[T1]{fontenc}  % Ensures proper rendering of characters in the output
\usepackage[utf8]{inputenc} % Ensures proper encoding of input files

\usepackage{microtype}    % Improves typography by adjusting spacing in small amounts

\usepackage{inconsolata}  % Improves monospaced font appearance

\usepackage{graphicx}     % Essential for including images

\usepackage{multirow}     % For tables with merged rows
\usepackage{booktabs}     % For professional-looking table rules (toprule, midrule, etc.)
\usepackage{array}        % For advanced tabular formatting and column types
\usepackage{arydshln}     % For dashed lines in tables

\usepackage{textgreek}    % For Greek characters in text mode

\usepackage{amsfonts}     % For mathematical fonts like \mathcal
\usepackage{amsmath}      % Core math package for advanced equation formatting
\usepackage{amssymb}      % Additional mathematical symbols
\usepackage{mathtools}    % Extends amsmath with additional features

\usepackage{algorithm}    % For algorithm environments
\usepackage{algorithmic}  % For content within algorithms (often used with algorithm)

\usepackage{listings}     % For code display, superior to verbatim environment
\usepackage{fancyvrb}     % For improved verbatim environments (alternative/complement to listings)

\usepackage{xcolor}       % For colored text and highlighting
\usepackage{lipsum}       % For generating placeholder text (remove for final paper)

\title{Multi-Agent Synergy-Driven Iterative Visual Narrative Synthesis}

\author{
  Wang Xi\textsuperscript{1,2,*} \quad Quan Shi\textsuperscript{3,*} \quad 
  Tian Yu\textsuperscript{1,2,\dag} \quad Yujie Peng\textsuperscript{1,2,\dag} \quad Jiayi Sun\textsuperscript{1,2\dag} \quad \\
  Mengxing Ren\textsuperscript{1,2} \quad
  Zenghui Ding\textsuperscript{1,\ddag} \quad Ningguang Yao\textsuperscript{1,\ddag}\\
  \textsuperscript{1}Hefei Institutes of Physical Science, Chinese Academy of Sciences\\
  \textsuperscript{2}University of Science and Technology of China\\
  \textsuperscript{3}Changzhou University\\
  xw\_cs@mail.ustc.edu.cn \quad s23040820006@smail.cczu.edu.cn \quad \\
  dingzenghui@iim.ac.cn \quad Yaong@mail.ustc.edu.cn
}

\begin{document}
\maketitle

\renewcommand{\thefootnote}{\fnsymbol{footnote}}
\footnotetext[1]{Equal contribution as first author}
\footnotetext[2]{Equal contribution as second author}
\footnotetext[3]{Corresponding author}
\renewcommand{\thefootnote}{\arabic{footnote}}

\begin{abstract}
Automated generation of high-quality media presentations is challenging, requiring robust content extraction, narrative planning, visual design, and overall quality optimization. Existing methods often produce presentations with logical inconsistencies and suboptimal layouts, thereby struggling to meet professional standards. To address these challenges, we introduce RCPS (Reflective Coherent Presentation Synthesis), a novel framework integrating three key components: (1) Deep Structured Narrative Planning; (2) Adaptive Layout Generation; (3) an Iterative Optimization Loop.  Additionally, we propose PREVAL, a preference-based evaluation framework employing rationale-enhanced multi-dimensional models to assess presentation quality across Content, Coherence, and Design. Experimental results demonstrate that RCPS significantly outperforms baseline methods across all quality dimensions, producing presentations that closely approximate human expert standards. PREVAL shows strong correlation with human judgments, validating it as a reliable automated tool for assessing presentation quality.
\end{abstract}

\section{Introduction}
\label{sec:introduction}
The automated generation of high-quality presentations (PPTs) is pivotal for efficient information dissemination, particularly in academic and business communication. Such presentations transform complex document information into clear and engaging visual narratives, yet their manual crafting is notoriously laborious and time-consuming \citep{Fu2022}. This process demands not only core information extraction and systematic organization but also the sophisticated design of visually compelling layouts, a skillset often requiring significant expertise.

Rapid advancements in Large Language Models (LLMs) \citep{OpenAI2023, Touvron2023, Templeton2024,Ouyang2022, Wei2022} have driven remarkable progress in automating complex tasks, including simulating human-like process handling \citep{Hong2023, Park2023}, rendering automated document-to-presentation synthesis seemingly feasible. However, despite this promise, a fundamental chasm remains: converting extensive documents into presentations that are simultaneously structurally coherent, visually appealing, and logically sound proves to be a formidable challenge. Existing LLM-driven approaches often falter, producing outputs with logical inconsistencies or suboptimal, non-adaptive layouts \citep{Bandyopadhyay2024, Author2025, Zheng2025}, thereby failing to meet professional standards.

The inherent limitations of these current methods underscore two persistent core challenges. Firstly, there is an insufficient capability for adaptive layout generation that is responsive to both content semantics and functional intent; template-based methods suffer from rigidity, while unconstrained LLM generation often disregards established design conventions. Secondly, achieving a high standard of holistic quality---encompassing coherence, content appropriateness, and visual design professionalism in a balanced manner---remains elusive. Existing approaches generally lack robust mechanisms for global narrative planning and the iterative, multi-modal refinement crucial for optimizing towards complex, multi-dimensional human preferences for overall presentation excellence.

To address these fundamental challenges, we introduce \textbf{RCPS} (Reflective Coherent Presentation Synthesis), a novel, integrated framework designed to emulate the human expert creation process. RCPS uniquely synergizes three critical capabilities:
\begin{enumerate}
    \item Deep Structured Narrative Planning via a Reflective Chain-of-Thought \textbf{(R-CoT)} to establish global coherence and logical flow from source documents;
    \item Content-and-Function Adaptive Layout Prototype Generation \textbf{(LPG)} to produce semantically appropriate and structurally sound initial visual arrangements;
    \item An Iterative Multi-Modal Optimization Loop for meticulous refinement. This holistic approach aims to produce presentations of significantly elevated quality, achieving a harmonious synthesis of content, structure, and design.
\end{enumerate}

To evaluate our framework, we designed a human-correlated evaluation framework named \textbf{PREVAL} and obtained the reliability of multi-modal optimization cycles through extensive experiments.

\section{Related Work}
\label{sec:related_work}

\noindent\textbf{Early Exploration: Rule-Based and Extractive Methods.}
Early attempts, dating back over two decades, primarily relied on heuristic rules and predefined templates to extract content for user-specified topics \citep{AlMasum2005, Winters2019}. Subsequent machine learning approaches improved sentence importance ranking and key phrase extraction \citep{HuWan2013, Wang2017, Sefid2019}. However, these methods were predominantly extractive. The generated slide content often consisted merely of aggregated original sentences, critically lacking the abstractive summarization and sophisticated information reorganization characteristic of authentic, human-crafted presentations \citep{Sun2021}. This fundamental limitation in narrative construction and content transformation highlighted the need for more advanced generative capabilities, a core motivation for the R-CoT planning module in RCPS.

\noindent\textbf{Advancements in Text Generation: Summarization and Sequence-to-Sequence Models.}
To generate more natural and concise slide text, research shifted towards framing presentation generation as a text summarization task, particularly Query-Based Single-Document Summarization (QSS) \citep{Sun2021}. While pioneering the integration of summarization, the D2S model's reliance on pre-existing or directly corresponding slide titles often proved impractical. Furthermore, its narrow focus on text summarization neglected visual layout, limiting practical utility. More recent multi-stage pipelines like DocPres \citep{Bandyopadhyay2024}, integrating LLMs and VLMs, aimed to decompose task complexity by including steps like outline derivation and image extraction. However, like many pipeline approaches\citep{Radford2021, Liu2021}, DocPres is susceptible to inter-stage error propagation and still faces significant challenges in ensuring global narrative coherence when integrating content from diverse document sections or hierarchical levels.

\noindent\textbf{Explicit Layout Prediction: Bridging Content and Visuals.}
The visual layout is undeniably critical. DOC2PPT \citep{Fu2022} represented a pioneering effort in end-to-end trainable models with explicit bounding box prediction for slide elements. Its primary drawback, however, was the stringent requirement for large-scale, fine-grained layout-annotated datasets, which are notoriously difficult and costly to acquire, thus hampering scalability and generalizability. Conversely, recent efforts employing fixed templates \citep{Zheng2025} ensure visual consistency but sacrifice the crucial adaptability of layout to varying content and functional intent. This tension between layout flexibility and data dependency motivates RCPS's LPG, which generates adaptive symbolic layout prototypes, aiming to strike a balance by deferring pixel-perfect rendering and avoiding the need for exhaustive coordinate-level annotations during initial generation.

\noindent\textbf{Leveraging Large Language Models and Agent-Based Systems.}
The advent of LLMs has opened new avenues \citep{Hong2023, Park2023}. Agent-based solutions are increasingly feasible \citep{fu-etal-2024-msi, xiong-etal-2024-watch}; for instance, the PPTC Benchmark \citep{Guo2024} evaluated LLM capabilities in executing multi-turn editing instructions within multi-modal environments, yet it also exposed their limitations in managing complex templates and performing robust spatial reasoning. Systems like PPTAgent \citep{Author2025} utilize LLM-driven agents for content generation and template population but often fall short in visual appeal and true layout flexibility. These studies collectively highlight a critical ongoing challenge: effectively balancing LLM-driven content generation with precise, adaptive, and aesthetically pleasing layout control \citep{shi-etal-2024-direct, lan-etal-2024-llm}. We address this through RCPS's synergistic

\begin{figure*}[t]
  \centering
  \includegraphics[width=0.97\textwidth]{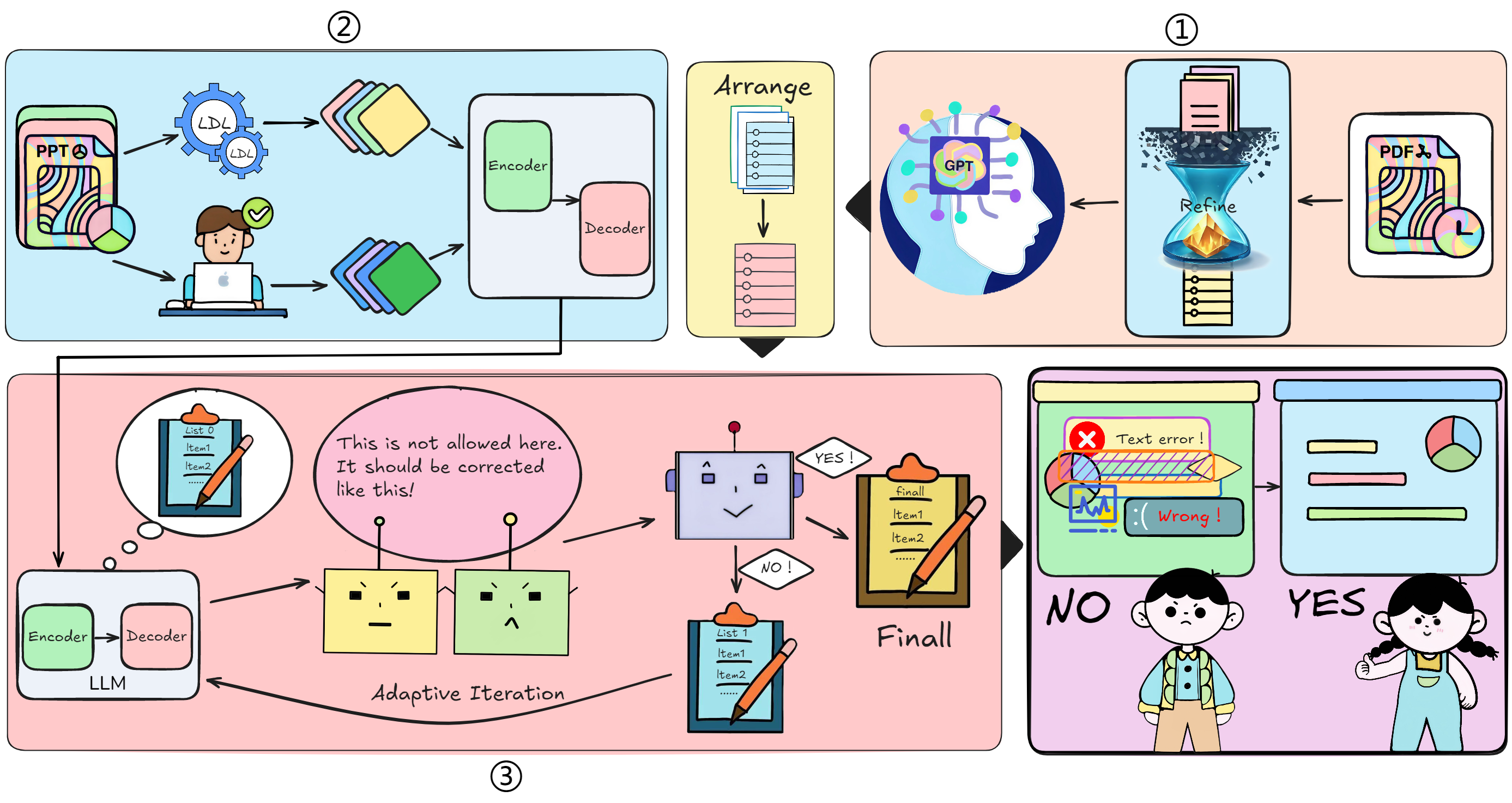}
  \caption{The RCPS framework, comprising three main components: (1) Reflective Chain-of-Thought (R-CoT) for structured narrative planning; (2) Layout Prototype Generator (LPG) for content-adaptive symbolic layout creation; and (3) Iterative Multi-Modal Optimization (IMR) Loop via multi-agent reflection for refining the presentation draft.}
  \label{fig:rcps-framework}
\end{figure*}

\noindent framework: R-CoT for content planning, LPG for adaptive initial layouts, and critically, an iterative multi-modal optimization loop that allows for fine-grained, feedback-driven refinement of both content and layout \citep{huang-etal-2024-towards, weng2025dothinkconformitylarge}, a mechanism often lacking in existing agent-based systems.

\section{Method}
\label{sec:method}
Automated generation of high-quality presentations is a complex challenge, requiring a synergistic integration of narrative planning, content adaptation, layout generation, and multi-dimensional quality optimization, moving beyond simple text processing. Existing methods often struggle with achieving global logical coherence \citep{Bandyopadhyay2024} and producing visually appropriate, adaptive designs \citep{Zheng2025}. To address these limitations, we propose RCPS, a multi-stage, iterative generation paradigm. As illustrated in Figure \ref{fig:rcps-framework}, RCPS uniquely combines: (1) R-CoT for structured narrative planning \citep{Wei2022, yu2025rethinkinggenerationhighqualitycot}; (2) LPG for content-adaptive layout prototyping, generating symbolic layout descriptions (LDL) by learning from high-quality examples \citep{Hu2024, ZAHEDIFAR2025104913}; (3) an iterative multi-modal optimization loop for fine-grained refinement using multi-agent reflection \citep{zhang2025aflowautomatingagenticworkflow, wang2025ragenunderstandingselfevolutionllm}.

\subsection{R-CoT}
\label{sec:rcot_method}
A presentation's narrative logic and fluency are foundational to its success, often necessitating intelligent information restructuring beyond the source document's original order. To achieve this, RCPS first employs an enhanced Reflective Chain-of-Thought mechanism. This process begins by parsing the input document \scalebox{0.9}{$D$} to extract primary content units \scalebox{0.9}{$u_{k} = (P_{k},F_{l})$} (text and figures) and their associated themes \scalebox{0.9}{$Theme_{k}$} via an LLM, forming a set of thematic units \scalebox{0.9}{$\mathcal{U} = \{(u_{k},Theme_{k})\}$}. The core of R-CoT then involves constructing an implicit Thematic Unit Graph \scalebox{0.9}{$\mathcal{G}_{T} = (\mathcal{U},\mathcal{E}_{T})$}, where edges \scalebox{0.9}{$\mathcal{E}_{T}$} (representing logical relations like 'support', 'contrast') are inferred by an LLM. Guided by R-CoT principles (details in Appendix A), a Planner Agent reasons over \scalebox{0.9}{$\mathcal{G}_{T}$} to generate a logically reordered narrative outline \scalebox{0.9}{$\mathcal{O}_{\text{narrative}}$} (e.g., \scalebox{0.9}{$[(\text{Stage}_{1}:\text{Background}), (\text{Stage}_{2}:\text{Core Results}), \ldots]$}). This outline ensures global coherence by capturing deep content logic beyond superficial sequential order.

Subsequently, each \scalebox{0.9}{$\text{Stage}_{i}$} is instantiated into a sequence of slide concepts \scalebox{0.9}{$\mathcal{O}_{\text{slides,i}} = \{ c_{i1}, \ldots, c_{iM_{i}}\}$}. Thematic units are assigned to the most appropriate concepts \scalebox{0.9}{$c_{ij}$}, each encapsulating a Key Message, source text \scalebox{0.9}{$P_{ij}$}, figures \scalebox{0.9}{$F_{ij}$}, and a functional type \scalebox{0.9}{$\text{type}_{j}$}. Finally, \scalebox{0.9}{$P_{ij}$} is refined by an LLM into concise bullet points \scalebox{0.9}{$T_{ij}$} suitable for presentation. The R-CoT stage thus provides a semantically rich and logically structured plan, including the content features \scalebox{0.9}{$feat_{content,j}$} and functional type \scalebox{0.9}{$type_j$} for each planned slide concept, which serve as input to the LPG.

\subsection{Adaptive Layout Prototype Generator}
\label{sec:lpg_method_revised} % 保持与您原文一致的标签，但内容是修改后的
To overcome the rigidity of fixed templates and the often arbitrary, low-quality layouts from general-purpose LLMs in unconstrained scenarios, RCPS introduces a specially designed and trained Layout Prototype Generator. Instead of producing pixel-perfect final layouts directly, LPG functions as a structured prior learning module. Its core objective is to transform abstract slide concepts (encoded with content features \scalebox{0.9}{$feat_{content,j}$} and functional type \scalebox{0.9}{$type_j$} from R-CoT) into content-adaptive, symbolically represented layout prototypes in the form of Layout Description Language (LDL) sequences, \scalebox{0.9}{$L^{(0)}$}. These prototypes, by learning from well-designed examples, inherently tend to adhere to basic design principles and offer high-quality starting points for subsequent iterative optimization.

\textit{Problem Formalization and Core Challenges.} Given a feature representation \scalebox{0.9}{$f$} of a slide concept, LPG aims to generate an LDL sequence \scalebox{0.9}{$L^{(0)} = (l_1, \ldots, l_M)$}. This sequence is learned by maximizing the likelihood of generating target sequences from a dataset of high-quality examples. The primary challenge is to effectively learn and express complex visual layout rules implicitly through this data-driven imitation process within a symbolic output space.

\textbf{Symbolic Layout Representation and Its Theoretical Motivation.} We opt for LPG to generate symbolic sequences following a LDL (detailed in Appendix \ref{appendix:ldl_vocabulary} ), rather than directly predicting continuous coordinate values. LDL uses predefined object vocabularies and attribute/positional tokens to describe layouts structurally. This choice is motivated by Information Theory and a Structural Focus (details in Appendix \ref{appendix:ldl}).

\textit{Model Architecture.} LPG's core employs a standard Transformer encoder-decoder architecture (specific configuration parameters in Appendix \ref{appendix:lpg_architecture_revised}), mapping input slide concept features \scalebox{0.9}{$f$} to context-aware representations. The decoder then autoregressively predicts each symbol \scalebox{0.9}{$l_t$} in the LDL sequence \scalebox{0.9}{$L^{(0)}$}.

\textit{Learning Objective: Imitation Learning with Standard Regularization.}
LPG's training objective focuses on imitating high-quality target LDL sequences (\scalebox{0.9}{$L_{\text{target}}$}) from a curated dataset (\scalebox{0.9}{$D_{\text{train}}$}), supplemented by standard L2 regularization. The objective function is:

\begin{equation}
\mathcal{L}_{\alpha} = -\sum_{(f,L_{\text{target}})\sim D_{\text{train}}} \left[ \log P_{\theta_{LPG}}(L_{\text{target}}|f) \right]
\end{equation}

\begin{equation}
\mathcal{L}_{\beta} = \alpha_{\text{L2}} \cdot \|\theta_{LPG}\|^2_2
\end{equation}

\begin{equation}
\mathcal{L}_{\text{obj}}(\theta_{LPG}) = \mathcal{L}_{\alpha} + \mathcal{L}_{\beta}
\end{equation}

Through exposure to well-designed \scalebox{0.9}{$L_{\text{target}}$} sequences, the model implicitly learns to generate prototypes that tend to follow established design conventions. (Further training details in Appendix \ref{appendix:lpg_implementation_revised}).

\textbf{Theoretical Advantages of LPG.}
LPG aims to: (1) Learn generalizable structural priors from data. (2) Provide robust symbolic starting points ($L^{(0)}$) that are then instantiated into an initial Structured Intermediate Representation (SIR) for subsequent refinement.

\subsection{Iterative Multi-Modal Optimization }
\label{sec:iterative_opt_revised} % 保持与您原文一致的标签，但内容是修改后的
While the symbolic layout prototypes \scalebox{0.9}{$L_{j}^{(0)}$} from LPG provide a strong starting point, achieving professional-quality presentations necessitates a dedicated refinement stage. To address this, RCPS employs an Iterative Multi-Modal Optimization (IMR) loop, emulating an expert's review-and-revise cycle. This appendix details the Adaptive iterations workflow algorithm.

The IMR loop begins by instantiating the LDL sequence \scalebox{0.9}{$L_{j}^{(0)}$} from LPG, along with content \scalebox{0.9}{$(T_{j},F_{j})$} from R-CoT, into an initial Structured Intermediate Representation (SIR), \scalebox{0.9}{$\text{SIR}_{j}^{(0)}$}. The SIR is a mutable, detailed representation of the slide draft, including element attributes for geometry, style, and content.

\begin{figure}[t]
  \begin{center}
    \includegraphics[width=1.15\columnwidth]{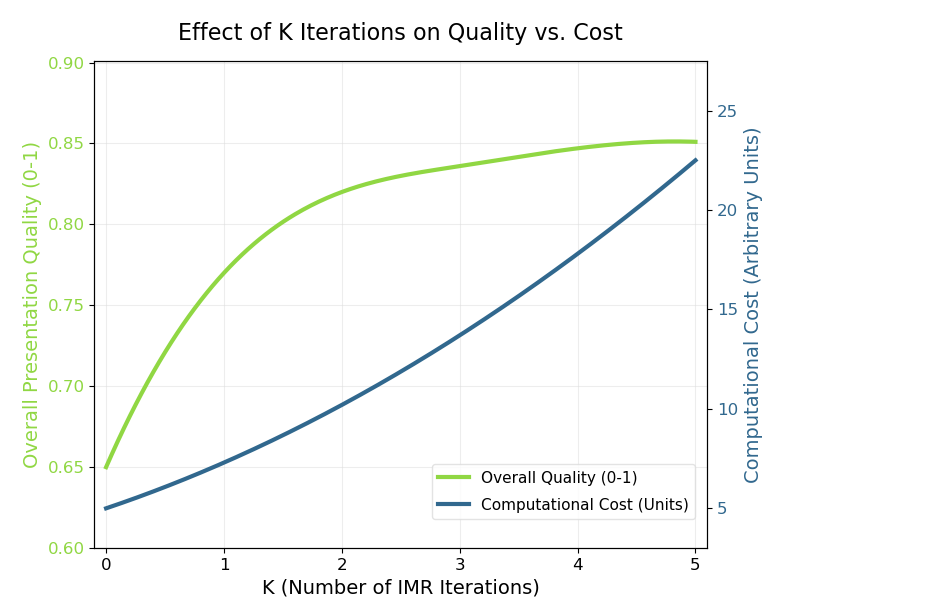}
    \caption{Effect of K Iterations on Quality vs. Cost}
    \label{fig:effect-k}
  \end{center}
\end{figure}

\textbf{The Core IMR Cycle (Adaptive iterations).} For each slide draft, represented by its SIR, \scalebox{0.9}{$\text{SIR}_{j}^{(t)}$} (initially \scalebox{0.9}{$t=0$}, the following operations are performed:

\textit{Visual Rendering.} The current $\text{SIR}_{j}^{(t)}$ is rendered into a visual preview image \scalebox{0.9}{$I_{j}^{(t)}$}, translating the SIR's structured data into a human-perceptible and machine-analyzable visual form.

\textbf{Structured Multi-Modal Critique Generation.} Two specialized critique modules analyze the rendered presentation:
    \begin{itemize}
        \item \textit{Visual Fidelity Critic (VLM-C):} A pre-trained VLM analyzes \scalebox{0.9}{$I_{j}^{(t)}$} and geometric/style attributes in $\text{SIR}_{j}^{(t)}$ to identify objective layout errors (e.g., Overlap, Misalignment, Text Overflow), outputting a structured list of issues \scalebox{0.9}{$\mathcal{C}_{\text{visual}}^{(t)}$} (see Appendix \ref{appendix:critique_format} for format).
        \item \textit{Logical Coherence Critic (LLM-C):} An LLM evaluates textual content within $\text{SIR}_{j}^{(t)}$ for clarity, conciseness, and coherence with the R-CoT plan, outputting \scalebox{0.9}{$\mathcal{C}_{\text{logic}}^{(t)}$}.
    \end{itemize}

\textbf{Reflective Editing via Planning and Parameterized Primitives.}
A Refinement Agent (LLM) processes the aggregated critique list \scalebox{0.9}{$\mathcal{C}_{j}^{(t)}$} using CoT reasoning and a predefined set of \textbf{parameterized Editing Primitives (EPs)} (Appendix \ref{appendix:editing_primitives} for examples). It first plans an ordered sequence of edits and then instantiates EPs with precise parameter values. These EPs directly and deterministically modify the attributes of the corresponding elements within the $\text{SIR}_{j}^{(t)}$.

\textbf{Draft Update and Termination.} The optimization process involves iteratively applying EP-generated modifications to the SIR, with each application of modifications transforming $\text{SIR}_{j}^{(t)}$  into $\text{SIR}_{j}^{(t + 1)}$. This iterative cycle continues until one of the predefined termination criteria is satisfied: specifically, when the severity of the critiques falls below a predetermined threshold, or when the process reaches a maximum allowable time limit \( T_{\text{max}} \). During these iterations, the system prioritizes addressing issues based on their severity ratings, with higher severity issues receiving precedence. To ensure that critical problems are promptly resolved, the optimization algorithm dynamically adjusts the priority of issues. If an issue has a high severity score, the system increases the optimization weight assigned to that issue, thereby prioritizing its resolution in subsequent iterations.

\section{Model Evaluation}
\label{sec:preval_new}
Accurate and comprehensive quality assessment of automatically generated presentations is fundamental to measuring and advancing the field. The limitations of traditional evaluation metrics are widely acknowledged, and relying solely on human ratings presents challenges in efficiency and consistency. To address this, we propose PREVAL (Preference-based Evaluation Framework via Learned Assessment), an evaluation framework designed to deeply emulate the holistic judgment of human experts. The core of PREVAL lies in utilizing multi-dimensional quality assessment models learned from human preferences. Crucially, PREVAL incorporates human-provided "rationales" to enhance the learning process, thereby training models that not only predict preferences but also possess stronger interpretability and sensitivity to features that humans deem important.

\subsection{Rationale-Enhanced Multi-dimensional Preference Model Learning}
PREVAL posits that the overall quality of a presentation can be viewed as a quality function across several key dimensions (e.g., Content, Coherence, Design). Since directly defining or computing this overall quality is intractable, PREVAL learns a human preference prediction model for each dimension by leveraging pairwise comparisons and their accompanying "rationales." This process comprises two main stages: an offline "Rationale-Enhanced Preference Model Learning" stage and an online "Model-Based Quality Assessment" stage.

The offline learning phase is central to PREVAL's evaluative capability. Its objective is to learn a set of scoring functions that can not only predict human preferences within specific dimensions but are also guided by the rationales provided by humans.

\textbf{Dataset Construction.}
We first construct a dataset comprising multiple presentation pairs $(\mathrm{PPT}_A, \mathrm{PPT}_B)$, human preference judgments for each predefined dimension (e.g., A is better than B, B is better than A, or A is comparable to B), and the rationales associated with these preferences. These rationales, composed of structured tags (e.g., text overflow, irrelevant image, logical clarity) and concise natural language explanations, serve as crucial supervisory signals.

\textbf{Multi-modal Feature Engineering and Representation.}
We define a function to map each presentation into a rich multi-modal feature vector. To implement this, we utilize powerful pre-trained large multi-modal models (LMMs) that process the textual content and slide images of presentations. These features capture textual semantics, high-level visual aesthetics, and structural properties. Additionally, we incorporate several interpretable handcrafted features (e.g., alignment scores, white space distribution, element counts) that can correspond to human-provided rationales.

\textbf{Attention-Based Multi-Task Learning.}
Our goal is to learn a scoring function for each dimension to predict its quality score. To achieve "rationale-enhancement" in learning, we propose an Attention-based Multi-Task Learning (AMTL) framework.

For each dimension, this framework has two primary objectives:
\begin{enumerate}
    \item Main Task: Preference Ranking.
    Given a pair of presentations (A, B) and their features $(x_A, x_B)$, the model learns to output two scores $(s_A, s_B)$ such that their difference aligns with human preferences. This task is optimized using a pairwise ranking loss function (e.g., logistic loss).

    \item Auxiliary Task: Rationale Prediction or Alignment.
    The model is concurrently trained to align with human-provided rationales. This includes:
    \begin{itemize}
        \item \textit{Rationale Attention Mechanism:} The model incorporates an attention mechanism whereby rationales (both structured tags and text explanations) dynamically influence the processing of presentation features, guiding the model to focus on the most relevant features as indicated by the rationales. For instance, if a rationale points to "text overflow," the model would pay more attention to features like text box fill rates or text density.
        \item \textit{Rationale Consistency Loss:} An auxiliary loss term is introduced. For example, if rationales are structured tags, a classification loss is used; if they are text embeddings, a cosine similarity loss is employed to align the model's internal "explanation" embeddings with those of human rationales.
    \end{itemize}
\end{enumerate}
The total loss function is a weighted sum of the main task loss and the auxiliary task losses, where the weights are tunable hyperparameters.

This AMTL approach ensures that the learned scoring functions not only predict human preferences but also make judgments based on the critical features explicitly highlighted by humans. The final output is a series of trained evaluation functions.

\subsection{PREVAL Evaluation Workflow}
Once the multi-dimensional evaluation functions are trained offline, PREVAL can be used to evaluate any new presentation online. This workflow efficiently outputs quantitative multi-dimensional quality scores and can optionally provide explanatory feedback.

The core workflow includes:
\begin{enumerate}
    \item \textit{Multi-modal Feature Extraction.}
    The input presentation is processed by the same feature extractors used during training to obtain its feature vector.

    \item \textit{Dimensional Quality Scoring.}
    The feature vector is fed into each trained scoring function, yielding a raw score, which is then normalized to a $[0, 1]$ interval.

    \item \textit{Overall Assessment and Explanation.}
    The dimensional scores form a quality profile, and a weighted aggregate score is computed. Furthermore, an explanation generation module can leverage the input features, dimensional scores, and outputs from the model's internal attention or rationale mechanisms to produce natural language feedback, reflecting the aspects PREVAL focused on during its evaluation.
\end{enumerate}

\section{Experiment} 
\label{sec:experiments}
This section provides a thorough evaluation of the proposed RCPS framework and the PREVAL assessment methodology. 

\subsection{Experimental Setup} 
\label{ssec:exp_setup}
\noindent\textit{Datasets.} The datasets employed in this study are as follows: (1) \textbf{RCPS Generation Dataset:} Including 1000 document-slide pairs from diverse

\begin{table*}[h!]
\centering
\scriptsize
\setlength{\tabcolsep}{7.5pt}
\renewcommand{\arraystretch}{0.95}
\begin{tabular}{lcccccccc}
\toprule
\multirow{2}{*}{\textbf{Method}} & \multicolumn{4}{c}{\textbf{PREVAL}} & \multicolumn{4}{c}{\textbf{Human}} \\
\cmidrule(lr){2-5} \cmidrule(lr){6-9}
 & Content & Coherence & Design & Overall & Content & Logic & Visual & Overall \\
\midrule
TextSum+T & 0.43 (±0.07) & 0.35 (±0.09) & 0.52 (±0.06) & 0.43 (±0.05) & 3.2 (±0.5) & 2.8 (±0.6) & 3.5 (±0.5) & 3.1 (±0.4) \\
DocPres & 0.58 (±0.06) & 0.47 (±0.08) & 0.49 (±0.07) & 0.51 (±0.05) & 4.1 (±0.4) & 3.7 (±0.5) & 3.8 (±0.4) & 3.9 (±0.3) \\
GPT-4o Zero-shot & 0.66 (±0.05) & 0.61 (±0.06) & 0.58 (±0.07) & 0.62 (±0.04) & 4.8 (±0.3) & 4.5 (±0.4) & 4.2 (±0.5) & 4.5 (±0.3) \\
GPT-4o + VisCoT Few-shot & 0.70 (±0.04) & 0.65 (±0.05) & 0.63 (±0.06) & 0.66 (±0.04) & 5.0 (±0.3) & 4.8 (±0.3) & 4.6 (±0.4) & 4.8 (±0.2) \\
\textbf{RCPS (Our method)} & \textbf{0.72 (±0.04)} & \textbf{0.73 (±0.05)} & \textbf{0.75 (±0.05)} & \textbf{0.73 (±0.03)} & \textbf{5.2 (±0.3)} & \textbf{5.4 (±0.2)} & \textbf{5.5 (±0.3)} & \textbf{5.4 (±0.2)} \\
\bottomrule
\end{tabular}
\caption{Main performance comparison. * $p < 0.01$ vs. strongest baseline. PREVAL [0,1]; Human 1-7 Likert.}
\label{tab:main_results}
\end{table*}

\noindent academic domains, specifically Computer Science (CS), Life Sciences (LS), and Social Sciences (SS), distributed in an 80:10:10 ratio.  (2) PREVAL Preference Dataset: 2000 pairwise PPT comparisons, annotated with dimensional preferences (Content, Coherence, Design) and structured rationales. Inter-Annotator Agreement (IAA) for preferences: Fleiss' Kappa = 0.78. (Further dataset curation details in Appendix \ref{app:Dataset Annotation}).

\noindent\textit{Evaluation Metrics.}
Primary: The PREVAL Framework reporting Content, Coherence, Design, and equally-weighted Overall scores (normalized via calibrated Sigmoid).
Human Evaluation: Five actresses evaluated 30 test documents (Content Relevance, Logical Flow [mapped to Coherence], Visual Appropriateness [mapped to Design],Overall Satisfaction; 7-point Likert). IAA: Krippendorff's \scalebox{0.9}{$\alpha=0.81$} (Details in Appendix \ref{app:human_eval_protocol_revised_en}).

\textit{Auxiliary Metrics:} ROUGE-L, Perplexity (PPL), Fréchet Inception Distance (FID), and Structural Edit Distance.

\noindent\textit{Baseline Methods.} (1) TextSum+Template; (2) DocPres \citep{Bandyopadhyay2024} (reproduced); (3) GPT-4o Zero-shot; (4) GPT-4o + VisCoT Few-shot. 

\noindent\textit{Statistical Analysis.} Key comparisons are supported by paired t-tests (\scalebox{0.9}{$p < 0.01$} indicating significance). Means and standard deviations (SD) are reported.

\subsection{RCPS Generation Performance} 
\label{ssec:rcps_performance}
RCPS consistently and significantly outperforms all baselines across PREVAL dimensions and human evaluations (Table \ref{tab:main_results}).

RCPS's advantages are particularly pronounced in Design (PREVAL: \textbf{0.75} vs. 0.63 for GPT-4o+VisCoT; Human-Visual: \textbf{5.5} vs. 4.6) and Coherence (PREVAL: \textbf{0.73} vs. 0.65; Human-Logic: \textbf{5.4} vs. 4.8). These results strongly support the efficacy of RCPS's LPG module and iterative multi-modal optimization for visual quality, and the R-CoT mechanism (Section \ref{sec:rcot_method}) for narrative coherence. RCPS achieves a superior, well-balanced performance across all dimensions.

\begin{table}[h]
\centering
\scriptsize
\setlength{\tabcolsep}{5pt}
\renewcommand{\arraystretch}{0.9}
\begin{tabular}{lcccc}
\toprule
\textbf{Method} & \textbf{R-L(↑)} & \textbf{PPL(↓)} & \textbf{FID(↓)} & \textbf{ED(↓)} \\
\midrule
 TextSum & 0.32 (±0.03) & 175.3 (±5.1) & 89.6 (±3.2) & 0.68 (±0.05) \\
    DocPres & 0.29 (±0.04) & 136.7 (±4.5) & 75.4 (±2.8) & 0.52 (±0.04) \\
    GPT-4o & 0.34 (±0.03) & 118.2 (±3.9) & 68.3 (±2.5) & 0.43 (±0.03) \\
    4o+VisCoT & 0.35 (±0.03) & 117.5 (±3.5) & \textbf{64.8} \textbf{(±2.7)} & 0.38 (±0.03) \\
    \textbf{RCPS} & \textbf{0.35 (±0.03)} & \textbf{102.7 (±3.1)} & 71.5 (±2.9) & \textbf{0.31 (±0.02)} \\
    \bottomrule
\end{tabular}
\caption{Auxiliary metrics with directional indicators. * $p < 0.01$ vs. best baseline.}
\label{tab:aux_metrics}
\end{table}

Auxiliary metrics (Table \ref{tab:aux_metrics}) show RCPS prioritizes abstractive refinement (PPL: \textbf{102.7}, best; ED: \textbf{0.31}, best) over verbatim extraction, with its FID (71.5) suggesting diverse, content-adaptive visual layouts.

\subsection{Ablation Studies} 
\label{ssec:ablation}
Ablation studies (Table \ref{tab:ablation_rcps}) confirm the critical contribution of each RCPS component.

\begin{table}[h]
\centering
% Change \scriptsize to a larger size
\small  % Options: \scriptsize < \footnotesize < \small < \normalsize
\setlength{\tabcolsep}{20pt}  % Increased from 6pt
\renewcommand{\arraystretch}{1.2}  % Adds more vertical space (1.0 is default)
\begin{tabular}{lc}
\toprule
\textbf{Method Variation} & \textbf{Overall} \\
\midrule
\textbf{RCPS (Full)} & \textbf{0.73} \\
RCPS w/o R-CoT Planning & 0.65* \\
RCPS w/o LPG (fixed template) & 0.65* \\
RCPS w/o Refinement (K=0) & 0.70* \\
\bottomrule
\end{tabular}
\caption{Ablation study. * $p < 0.01$ drop vs. Full RCPS.}
\label{tab:ablation_rcps}
\end{table}

Removing R-CoT most significantly impacted Coherence (absolute drop of 0.15 in PREVAL-Coherence score), underscoring its role in narrative planning. Replacing LPG with a fixed template severely degraded Design (drop of 0.20). Disabling iterative refinement substantially reduced Design (drop of 0.09).

\subsection{PREVAL Framework Validation} 
\label{ssec:preval_validation}
\begin{figure}[t]
  \centering
  \includegraphics[width=\columnwidth]{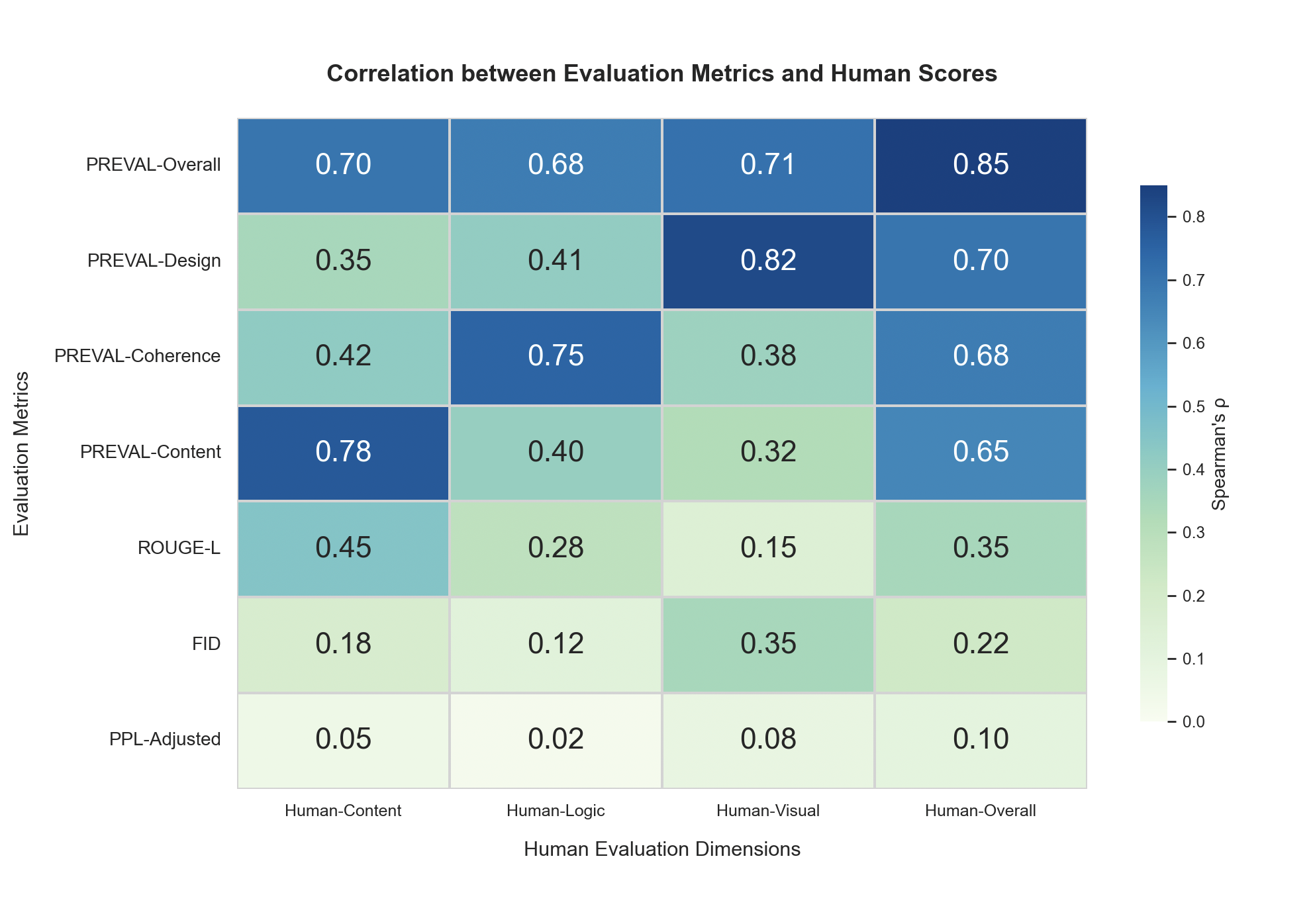}
  \caption{Correlation between PREVAL scores and human judgments (Spearman's \scalebox{0.9}{$\rho=0.85$} for Overall scores).}
  \label{fig:correlation_heatmap_preval}
\end{figure}

PREVAL's reliability is validated by its strong correlation with human judgments (Spearman's \scalebox{0.9}{$\rho=0.85$} for Overall scores, \scalebox{0.9}{$p < 0.001$}; Figure \ref{fig:correlation_heatmap_preval}). Kendall's \scalebox{0.9}{$\tau$} averaged 0.71 for dimensional rank agreement. Crucially, PREVAL captures quality dimensions missed by traditional metrics. For instance, TextSum+Template (acceptable ROUGE-L) receives low PREVAL-Coherence/Design scores, accurately identifying its flaws. A systematic analysis on a curated defect-set  shows PREVAL achieves a significantly higher F1-score (0.82 vs. 0.45 for ROUGE-L) in identifying problematic presentations.

\section{Conclusion}
\label{sec:conclusion}
We have proved the remarkable advantages of the RCPS framework in automatically generating high-quality presentations. By combining reflective thinking chain with structured planning, content-function adaptive layout generation and multi-agent iterative optimization, RCPS can produce presentations that are superior to the existing baseline methods in content, logic and design. At the same time, the PREVAL evaluation framework has also been verified as a reliable and effective evaluation tool, and its scoring results are highly related to human judgment, and can provide more comprehensive and in-depth quality insight than traditional indicators.

Despite the remarkable progress, there are still limitations in our work. The performance of RCPS depends on the ability of LMM/VLM to some extent, especially the understanding of complex document structure and subtle aesthetic judgment. The generalization ability of layout prototype generator still has room for improvement for unseen slide types or extreme content (super-long text and super-many pictures). Although the PREVAL framework is powerful, the training of its preference model needs high-quality human annotation data, and the current "causal perception" is still preliminary, which fails to achieve strict causal inference.

\section*{Limitations}
This paper presents significant advancements in automated presentation generation, but several limitations should be acknowledged: 

(1)Our approach is heavily reliant on the capabilities of foundation models (LLMs and VLMs), thereby inheriting their limitations in handling extremely technical content, complex document structures, and domain-specific terminology. This restricts the system's adaptability to highly specialized contexts. 

(2)Although RCPS demonstrates strong performance across the tested domains, its generalization to highly specialized fields. This gap may hinder its applicability in niche areas where domain expertise is critical.

(3)Our Layout Prototype Generator, while adaptive, still struggles with extremely unconventional slide compositions or highly specialized visualization types. This limitation may affect the system's ability to produce presentations with unique or highly creative designs.

(4)The full RCPS pipeline, particularly the iterative optimization phase, requires substantial computational resources. This demand may limit its practical deployment in resource-constrained environments, such as small businesses or educational institutions with limited access to high-performance computing. 

(5)The PREVAL framework, despite its strong correlation with human judgments, relies heavily on extensive human-annotated preference data for training. Scaling this requirement across all potential domains and presentation styles may be challenging and resource-intensive. 

(6) While our evaluation is comprehensive, it primarily focuses on English-language presentations with Western design conventions. The cross-cultural and multilingual aspects of presentation quality are underexplored, warranting further investigation to ensure the system's global applicability.

\section*{Acknowledgments}
We would like to express our sincere gratitude to the anonymous reviewers for their insightful feedback and constructive suggestions, which significantly contributed to the improvement of this paper. We are also deeply appreciative of the annotators who participated in our studies. Their efforts were instrumental in shaping the research outcomes.

\bibliography{latex/main}

\newpage

\appendix

\section{Prompt Engineering for R-CoT: Implementing Deep Structured Narrative Planning}
\label{appendix:rcot_prompt}

This appendix details the prompt engineering implementation of the Reflective Chain-of-Thought (R-CoT) module within our RCPS framework. R-CoT utilizes a series of structured prompts for GPT-4 to extract information, plan the narrative structure, and generate initial slide concepts from a source document. The goal of R-CoT is to provide a logically coherent and content-rich starting point for the subsequent Adaptive Layout Generation (LPG) and Iterative Multi-Modal Optimization (IMR) stages. All prompts are designed for GPT-4 and have undergone multiple rounds of iterative validation to ensure robustness. A general error handling mechanism involving a retry with a simplified prompt is applied if an LLM fails to produce a valid JSON output; persistent failures are logged.

\subsection{Stage 1: Document Parsing and Semantic Unit Annotation}

\textbf{Objective:} To parse the source document (pre-processed into Markdown format) into content units and annotate them with initial semantic information, including figure/table references.

\textbf{Input:} Source document text in Markdown format.

\textbf{Implementation Steps \& Core Prompt Elements:}
\begin{enumerate}
\item \textbf{Initial Parsing, Visual/Table Reference Extraction, and Placeholder Creation:}

The Markdown document is parsed using the Python `mistune` library. For images (`![alt](path)`) and tables represented via specific Markdown extensions (e.g., `Table: [Caption]` followed by `[Markdown Table]`), their paths/IDs and descriptive text (alt text or caption) are extracted. Image files are assigned unique IDs (e.g., `doc\_img\_001`) based on their order of appearance, and a mapping table from these IDs to file paths is established. Table content is converted into concise text summaries using a dedicated LLM prompt focused on extracting key data points and trends. Crucially, before passing segments to the main semantic unit identification LLM, complex Markdown for images and tables is replaced with simplified placeholders incorporating their extracted textual descriptions/summaries and IDs (e.g., `[IMAGE\_DESCRIPTION: doc\_img\_001: Diagram flusso di lavoro.]` or `[TABLE\_SUMMARY: doc\_table\_001: Risultati principali mostrano un aumento del 20\%.]`). This simplifies the input for subsequent LLM processing.

\item \textbf{LLM Semantic Unit Identification \& Annotation}

\textit{System Message:} "You are a precise document semantic segmenter. You will identify and characterize all distinct semantic content units from the provided Markdown segment."

\textit{User Prompt:}

\begin{lstlisting}[caption=User Prompt for Semantic Unit Annotation, basicstyle=\ttfamily\footnotesize, breaklines=true]
Objective: For the provided Markdown document segment (which has image/table raw data replaced by their textual description/summary placeholders), identify and characterize all distinct semantic content units.

Instructions:

Segment the text into the smallest meaningful units. These include headings (of various levels), paragraphs, list items, and the textual descriptions/summaries of images and tables (now represented as placeholders).

For each unit, assign:
a. unit_id: A unique identifier (e.g., "doc_unit_001").
b. text_content: The full text of the unit (for placeholders, this is their descriptive text).
c. unit_type: Categorize from [heading_1, heading_2, heading_3, paragraph, list_item, image_description_placeholder, table_summary_placeholder, code_block, blockquote].
d. concise_theme: A 3-5 word theme summarizing the unit's core topic.
e. source_visual_id: If unit_type is 'image_description_placeholder' or 'table_summary_placeholder', provide the corresponding pre-extracted ID (e.g., "doc_img_001", "doc_table_001"). Default to null for other types.
Output: A JSON list of these unit objects.
Input: \texttt{{markdown_segment_with_
visual_text_placeholders}}
\end{lstlisting}
\end{enumerate}

\textbf{Image/Table Data Linkage:} The source\_visual\_id (e.g., "doc\_img\_001") is used in subsequent stages to associate with externally stored image files or structured table data. The mapping table and placeholder strategy established in step 1 ensure that the LPG module can access the corresponding visual resources for feature extraction via these IDs.

\subsection{Stage 2: Theme-Driven Narrative Module Construction and Logical Ordering}

\textbf{Objective:} To organize content units into logically coherent macro-narrative modules based on their themes.

\textbf{Input:} List of all content units (including their unit\_id, text\_content, and concise\_theme) from Stage 1.

\textbf{Implementation Steps \& Core Prompt Elements:}
\begin{enumerate}
\item \textbf{Thematic Embedding and Clustering:}

Embeddings for the `text\_content` of each unit are generated using Sentence-BERT (all-MiniLM-L6-v2). DBSCAN clustering algorithm is applied to group units by thematic similarity. Its `eps` parameter is dynamically adjusted within the range [0.2, 0.4] based on the total number of content units in the document (e.g., using a linear scaling factor: $\text{eps} = 0.2 + 0.2 \times (\text{num\_units} / \text{MAX\_UNITS\_THRESHOLD})$, capped at 0.4), to form thematic clusters. A representative theme for each cluster is generated by selecting the most frequent `concise\_theme` among its member units, or by an LLM summarizing the cluster's content if themes are too diverse.

\item \textbf{LLM Narrative Module Construction \& Ordering}

\textit{System Message:} "You are an expert in structuring complex information into a compelling narrative flow for presentations. Your task is to group thematic clusters into logical narrative modules and order them effectively."

\textit{User Prompt:}

\begin{lstlisting}[caption=User Prompt for Narrative Module Construction, basicstyle=\ttfamily\footnotesize, breaklines=true]
Objective: Given themed content unit clusters (each with a representative theme and member unit_ids), group them into 3-6 ordered "Narrative Modules". Define each module's role in the overall presentation narrative.

Instructions (Chain-of-Thought & Reflection):

Initial Module Proposal (Thought): Review the input clusters and their representative themes. Propose an initial set of Narrative Modules by grouping semantically related clusters. For each proposed module, assign a tentative descriptive module_name and list its member_cluster_ids.

Logical Sequencing & Role Definition (Thought): Determine the optimal presentation order for these proposed modules. For each ordered module, define its module_role from a predefined set (e.g., "Introduction/Context", "Problem Statement", "Proposed Method/Solution", "Experimental Setup", "Results & Analysis", "Discussion", "Conclusion & Future Work"). Justify your ordering based on logical progression (e.g., "Module A (Problem) must precede Module B (Solution)"). Ensure a clear narrative arc.

Coherence & Completeness Review (Reflection):
a. Is the sequence of modules logically sound and easy to follow? Does it tell a coherent story?
b. Are there any significant thematic gaps or redundancies between modules?
c. Could the grouping of clusters into modules be improved for better thematic cohesion or narrative impact?
Based on this review, provide the final, refined list of ordered Narrative Modules. Each module object in the output JSON list must contain: module_id (unique), module_name, module_role, and member_cluster_ids. If your final proposal differs significantly from an initial implicit thought process due to reflection, briefly state the key reasoning for the change.
    Output: A JSON list of ordered Narrative Module objects.
\end{lstlisting}
\end{enumerate}

\textbf{Error Handling Note:} As mentioned in the section introduction, if the LLM outputs an invalid JSON format for this stage, a retry mechanism is employed.

\subsection{Stage 3: Presentation Outline Generation and Slide Structure Planning}

\textbf{Objective:} To map narrative modules to a standard presentation outline and plan the slide structure (number of slides and key content points) for each stage of the outline.

\textbf{Input:} Ordered list of "Narrative Modules" (each with module\_id, module\_name, module\_role) from Stage 2.

\textbf{LLM Core Instructions:}

\textit{System Message:} "You are a strategic presentation architect. Your task is to translate high-level narrative modules into a concrete presentation outline and plan the distribution of content across slides."

\textit{User Prompt:}

\begin{lstlisting}[caption=User Prompt for Outline Generation, basicstyle=\ttfamily\footnotesize, breaklines=true]
Objective: Convert the ordered list of Narrative Modules into a standard presentation outline consisting of 4-7 logical stages. For each stage, plan the slide allocation and identify key content points.

Instructions (Chain-of-Thought & Reflection):

Stage Mapping (Thought): Review the input Narrative Modules and their roles. Group adjacent or related modules to form logical presentation stages (e.g., "1. Introduction", "2. Methodology", "3. Results", "4. Discussion", "5. Conclusion"). Justify non-obvious mappings or groupings. A single Narrative Module might map to a stage, or multiple related modules might be grouped into one stage.

Slide Allocation Planning (Thought): For each defined stage, considering the volume and importance of its source Narrative Module(s), propose an allocated_slide_count. This should generally be an integer between 1 and 3 slides per major sub-theme or key concept within the stage, with total stage slides typically ranging from 1-5. The LLM should infer these major sub-themes/concepts from the module_name and module_role of the source modules.

Key Content Points Identification (Thought): For each stage, list 2-4 distinct key_content_points that must be covered across its allocated slides. These points should be derived from the core messages of the source Narrative Module(s).

Outline Validation (Reflection): Review the generated outline:
a. Does the stage progression ensure comprehensive coverage of all input Narrative Modules?
b. Is the allocated_slide_count for each stage proportionate to its content volume and narrative importance?
c. Are the key_content_points representative, sufficient, and distinct for each stage?
Provide the final presentation outline. Each stage object in the output JSON list must include: stage_number (integer), stage_title (e.g., "1. Introduction"), source_module_ids (list of module_ids contributing to this stage), allocated_slide_count, and key_content_points (list of strings). If adjustments were made during reflection, note the change and reason.
Output: A JSON list of outline stage objects.
\end{lstlisting}

\subsection{Stage 4: Slide Concept Instantiation and Content Refinement}

\textbf{Objective:} To generate concrete slide concepts for each slide planned in an outline stage, and refine source text into concise bullet points for presentation.

\textbf{Input:}
\begin{itemize}
\item A single outline stage object (from Stage 3), which includes allocated\_slide\_count, key\_content\_points, and source\_module\_ids.
\item All original content units (from Stage 1) that belong to the source\_module\_ids of the input outline stage.
\end{itemize}

\textbf{LLM Core Instructions:}

\textit{System Message:} "You are an efficient slide crafter, adept at transforming source material into impactful presentation content. You will generate distinct slide concepts and refine text into clear bullet points."

\textit{User Prompt:}

\begin{lstlisting}[caption=User Prompt for Slide Concept Instantiation, basicstyle=\ttfamily\footnotesize, breaklines=true]
Objective: For the input presentation stage (details provided below), generate exactly {{allocated_slide_count}} distinct slide concepts. Ensure that the key_content_points for this stage are reasonably distributed and covered across these generated slide concepts.

Instructions:

Strictly generate {{allocated_slide_count}} slide concepts. Each concept should aim to cover one or more related key_content_points from the input stage, or aspects thereof.

For each slide concept, define the following:
a. slide_title: Create a concise and informative title (max 8 words) reflecting the content of this specific slide.
b. key_message: Formulate a single, impactful sentence (max 20 words) summarizing the main takeaway of this slide.
c. functional_type: Select ONE from the PREDEFINED list of 10 types: ["title_main", "agenda", "section_header", "content_text_only", "content_text_image_left", "content_text_image_right", "content_image_only", "comparison_table", "key_takeaways", "thank_you_contact"]. Choose the type that best suits the intended content and visual elements for this slide.
d. source_unit_ids: List the unit_id(s) from the Input Content Units for this Stage (provided below) that are the primary sources of information for this specific slide concept. Ensure that all relevant unit_ids constituting the content of the overall input stage are reasonably distributed across the {{allocated_slide_count}} slide concepts.
e. bullet_points: Based on the text_content of the source_unit_ids assigned to this slide, generate 3-4 concise bullet points. Each bullet point should be 7-12 words long and clearly convey a key piece of information.
f. primary_visual_id: If one specific visual (image or table summary placeholder, identified by its source_visual_id from the source_unit_ids) is central to this slide's message, specify its ID (e.g., "doc_img_001"). Otherwise, set to null.
Output: A JSON list containing {{allocated_slide_count}} slide concept objects.

Input Stage Details: {{single_outline_stage_object_from_stage_3}}
Input Content Units for this Stage (a list of unit objects from Stage 1, filtered by source_module_ids):
{{list_of_relevant_units_from_stage_1_for_this_stage}}

[A concise Few-Shot Example is provided here, demonstrating the transformation of one input stage (with 2 allocated slides) into the correct JSON output format for two slide concepts, including how source_unit_ids are selected and bullet_points are generated. This example is available in the supplementary material / code repository.]
\end{lstlisting}

\textbf{LPG Input Linkage:} The bullet\_points ($T_{ij}$), functional\_type ($type_j$), and images (via primary\_visual\_id which links to $F_{ij}$) from this stage form the core source for the LPG's input features $feat_{content,j}$ and $type_j$.

\section{Layout Description Language (LDL) for Adaptive Presentation Synthesis}
\label{appendix:ldl}

The Layout Description Language (LDL) is a core component of our RCPS framework, enabling the Layout Prototype Generator (LPG) to produce structured, symbolic representations of slide layouts. This appendix details the vocabulary and design principles of LDL. The primary goal of LDL is to provide a concise yet expressive way to define the macro-structure and key characteristics of a slide layout, serving as a strong initial prior for subsequent multi-modal optimization. It focuses on element types, their semantic attributes, and their general placement, rather than pixel-perfect coordinates or complex inter-element relational constraints, which are refined in later stages.

\subsection{LDL Vocabulary}
\label{appendix:ldl_vocabulary}

The LDL vocabulary is organized into several categories:

\subsubsection{Slide Type Tokens}
\label{appendix:ldl_slide_types}
These tokens define the overall template or purpose of the slide.
\begin{itemize}
    \item \texttt{SLIDE\_TITLE}: For a main title slide.
    \item \texttt{SLIDE\_CONTENT\_SINGLE\_COL}: For content arranged in a single column.
    \item \texttt{SLIDE\_CONTENT\_TWO\_COL}: For content arranged in two columns.
    \item \texttt{SLIDE\_SECTION\_HEADER}: For a slide introducing a new section.
    \item \texttt{SLIDE\_IMAGE\_CAPTION}: For a slide primarily featuring an image with a caption, typically with the image as the dominant element and a smaller text block for the caption.
    \item \texttt{SLIDE\_BLANK}: For a blank slide, often used for transitions or full-slide visuals.
\end{itemize}

\subsubsection{Element Type Tokens}
\label{appendix:ldl_element_types}
These tokens specify the type of content element to be placed on the slide.
\begin{itemize}
    \item \texttt{ELEM\_TITLE}: A primary title or heading for the slide content.
    \item \texttt{ELEM\_SUBTITLE}: A secondary title or subheading.
    \item \texttt{ELEM\_TEXT\_BODY}: A block of text, typically bullet points or paragraphs.
    \item \texttt{ELEM\_IMAGE}: A placeholder for an image.
    \item \texttt{ELEM\_CHART}: A placeholder for a chart or graph.
    \item \texttt{ELEM\_TABLE}: A placeholder for a table.
    \item \texttt{ELEM\_FOOTER}: A footer element, often containing page numbers or disclaimers.
    \item \texttt{ELEM\_HEADER}: A header element, typically at the top of the slide.
\end{itemize}

\subsubsection{Element Attribute Tokens}
\label{appendix:ldl_element_attributes}
These tokens describe semantic or structural characteristics of an element's content, guiding layout adaptation during instantiation and subsequent optimization.
\begin{itemize}
    \item \texttt{ATTR\_TEXT\_POINTS\_FEW}: Text body contains a small number of bullet points (e.g., 1-3).
    \item \texttt{ATTR\_TEXT\_POINTS\_MEDIUM}: Text body contains a moderate number of bullet points (e.g., 4-6).
    \item \texttt{ATTR\_TEXT\_POINTS\_MANY}: Text body contains many bullet points (e.g., >6).
    \item \texttt{ATTR\_TEXT\_LENGTH\_SHORT}: Text content is concise.
    \item \texttt{ATTR\_TEXT\_LENGTH\_LONG}: Text content is extensive.
    \item \texttt{ATTR\_IMAGE\_ASPECT\_WIDE}: Image has a landscape aspect ratio.
    \item \texttt{ATTR\_IMAGE\_ASPECT\_SQUARE}: Image has a roughly square aspect ratio.
    \item \texttt{ATTR\_IMAGE\_ASPECT\_TALL}: Image has a portrait aspect ratio.
    \item \texttt{ATTR\_SIZE\_PRIMARY}: Element is of primary importance/visual weight and should occupy a significant area within its assigned zone.
    \item \texttt{ATTR\_SIZE\_SECONDARY}: Element is of secondary importance/visual weight and may occupy a smaller area.
    \item \texttt{ATTR\_CONTENT\_DENSE}: Indicates the element contains dense information (e.g., a complex table or detailed diagram), potentially requiring more space or careful layout.
    \item \texttt{ATTR\_CONTENT\_SPARSE}: Indicates the element contains sparse information, allowing for more generous spacing.
\end{itemize}

\subsubsection{Position Tokens}
\label{appendix:ldl_position_tokens}
These tokens define the general placement zone or alignment for an element on the slide. Multiple position tokens can often be combined to specify a more precise location (e.g., \texttt{POS\_TOP} and \texttt{POS\_CENTER} together suggest top-center placement, as illustrated in Section \ref{appendix:ldl_example}). These tokens guide the initial instantiation of the layout by the LDL Instantiator.
\begin{itemize}
    \item \texttt{POS\_TOP}: Element is placed in the top region of the slide or its parent container/zone.
    \item \texttt{POS\_MIDDLE}: Element is placed in the middle region (vertically) of the slide or its parent container/zone.
    \item \texttt{POS\_BOTTOM}: Element is placed in the bottom region of the slide or its parent container/zone.
    \item \texttt{POS\_LEFT}: Element is placed in the left region/column of the slide or its parent container/zone.
    \item \texttt{POS\_CENTER}: Element is placed in the center region (horizontally) of the slide or its parent container/zone.
    \item \texttt{POS\_RIGHT}: Element is placed in the right region/column of the slide or its parent container/zone.
    \item \texttt{POS\_FULL\_WIDTH}: Element spans the full width of its available content area or zone.
    \item \texttt{POS\_HALF\_WIDTH\_LEFT}: Element occupies the left half of a two-column layout or a similar designated area.
    \item \texttt{POS\_HALF\_WIDTH\_RIGHT}: Element occupies the right half of a two-column layout or a similar designated area.
    \item \texttt{POS\_TOP\_LEFT}, \texttt{POS\_TOP\_RIGHT}, \texttt{POS\_BOTTOM\_LEFT}, \texttt{POS\_BOTTOM\_RIGHT}: For general corner placements within a relevant zone.
\end{itemize}

\subsubsection{Special Sequence Tokens}
\label{appendix:ldl_special_tokens}
\begin{itemize}
    \item \texttt{< SOS >}: Start of Sequence. Marks the beginning of an LDL description for a slide.
    \item \texttt{<EOS>}: End of Sequence. Marks the end of an LDL description.
    \item \texttt{<SEP>}: Separator. Separates the description of one element from the next within the LDL sequence.
\end{itemize}

\subsection{Example of an LDL Sequence and Its Interpretation}
\label{appendix:ldl_example}

Below is an example of an LDL sequence that the LPG might generate for a two-column content slide.

\begin{figure}[h]
  \centering
  \includegraphics[width=\columnwidth]{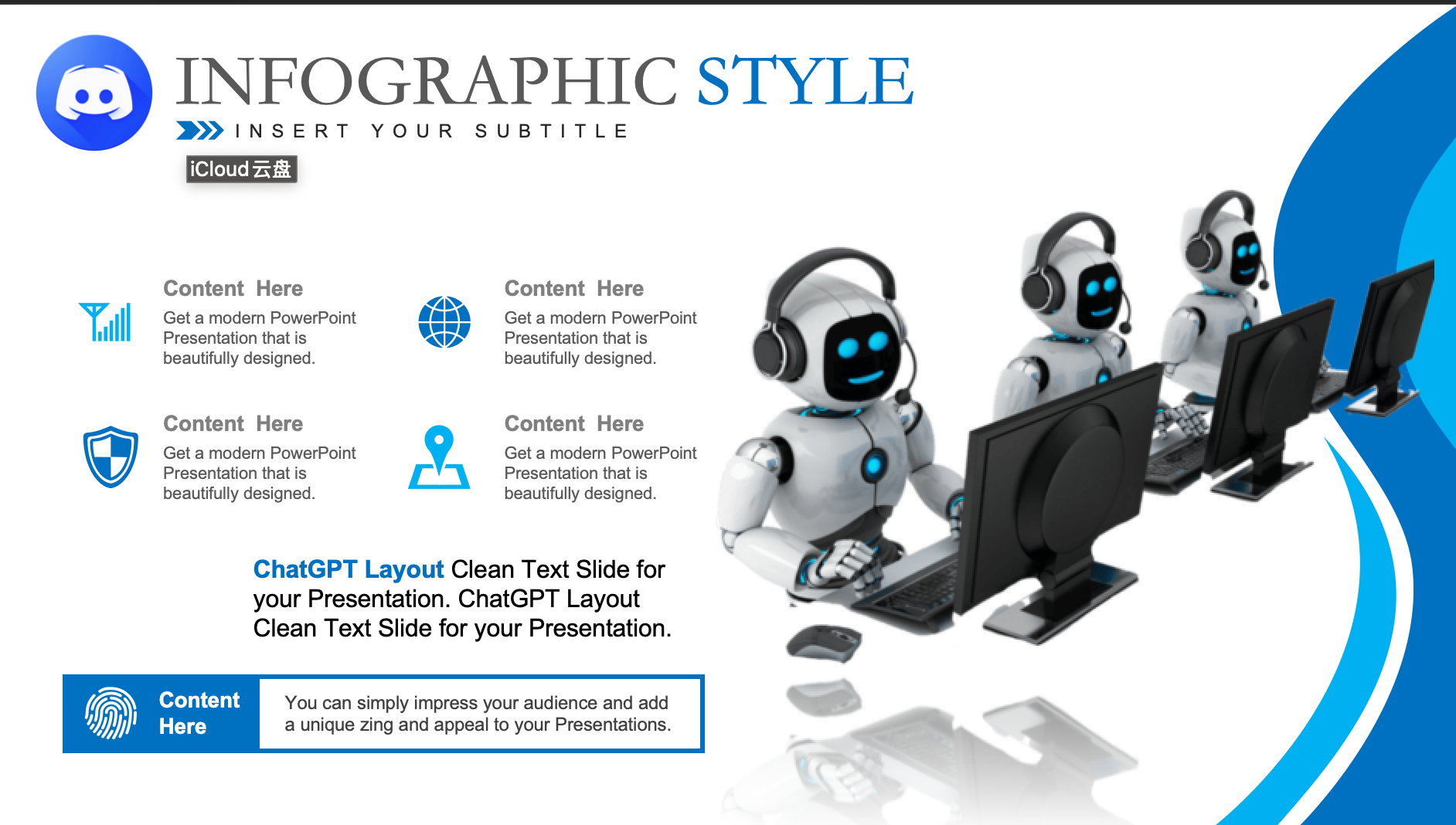}
  \caption{A PowerPoint presentation on education in terminology}
  \label{fig:Presentation_PowerPoint}
\end{figure}

{ % 开始一个局部作用域
\small % 设置字体大小
\begin{lstlisting}[caption=Example LDL for Infographic Style Slide, basicstyle=\ttfamily\footnotesize, breaklines=true, escapechar=|]
<SOS>
SLIDE_TITLE ATTR_STYLE_MODERN_INFOGRAPHIC <SEP> /* Assuming a new attribute for overall style */

/* Header Section */
ELEM_IMAGE ATTR_IMAGE_ASPECT_SQUARE ATTR_SIZE_SECONDARY POS_TOP_LEFT <SEP> /* Discord Logo */
ELEM_TITLE ATTR_TEXT_LENGTH_SHORT ATTR_SIZE_PRIMARY POS_TOP POS_CENTER <SEP> /* INFOGRAPHIC STYLE */
ELEM_SUBTITLE ATTR_TEXT_LENGTH_MEDIUM POS_TOP POS_CENTER <SEP> /* INSERT YOUR SUBTITLE - below title */
ELEM_TEXT_BODY ATTR_TEXT_LENGTH_SHORT POS_TOP_LEFT ATTR_STYLE_TAG <SEP> 

/* Main Content - Left Column (approximated as two grouped content blocks) */
/* Block 1 & 2 (Top-Left Quadrant) */
ELEM_CONTENT_BLOCK ATTR_LAYOUT_ICON_LEFT ATTR_TEXT_POINTS_FEW ATTR_TEXT_LENGTH_SHORT POS_MIDDLE_LEFT_UPPER <SEP>
ELEM_CONTENT_BLOCK ATTR_LAYOUT_ICON_LEFT ATTR_TEXT_POINTS_FEW ATTR_TEXT_LENGTH_SHORT POS_MIDDLE_LEFT_UPPER <SEP> 
/* Block 3 & 4 (Bottom-Left Quadrant) */
ELEM_CONTENT_BLOCK ATTR_LAYOUT_ICON_LEFT ATTR_TEXT_POINTS_FEW ATTR_TEXT_LENGTH_SHORT POS_MIDDLE_LEFT_LOWER <SEP>
ELEM_CONTENT_BLOCK ATTR_LAYOUT_ICON_LEFT ATTR_TEXT_POINTS_FEW ATTR_TEXT_LENGTH_SHORT POS_MIDDLE_LEFT_LOWER <SEP>

/* Main Content - Right Column */
ELEM_IMAGE ATTR_IMAGE_ASPECT_WIDE ATTR_SIZE_PRIMARY POS_MIDDLE_RIGHT <SEP> /* Robots Image */

/* Middle-Bottom Text */
ELEM_TEXT_BODY ATTR_TEXT_LENGTH_LONG POS_CENTER_HORIZONTAL POS_BOTTOM_MIDDLE_SECTION <SEP> /* ChatGPT Layout text */

/* Footer Section */
ELEM_FOOTER_FEATURED ATTR_LAYOUT_ICON_LEFT ATTR_TEXT_LENGTH_MEDIUM POS_BOTTOM POS_FULL_WIDTH <SEP> /* Bottom bar with icon and text */

<EOS>
\end{lstlisting}
} % 结束局部作用域，字体大小恢复

\begin{figure}[h]
  \centering
  \includegraphics[width=\columnwidth]{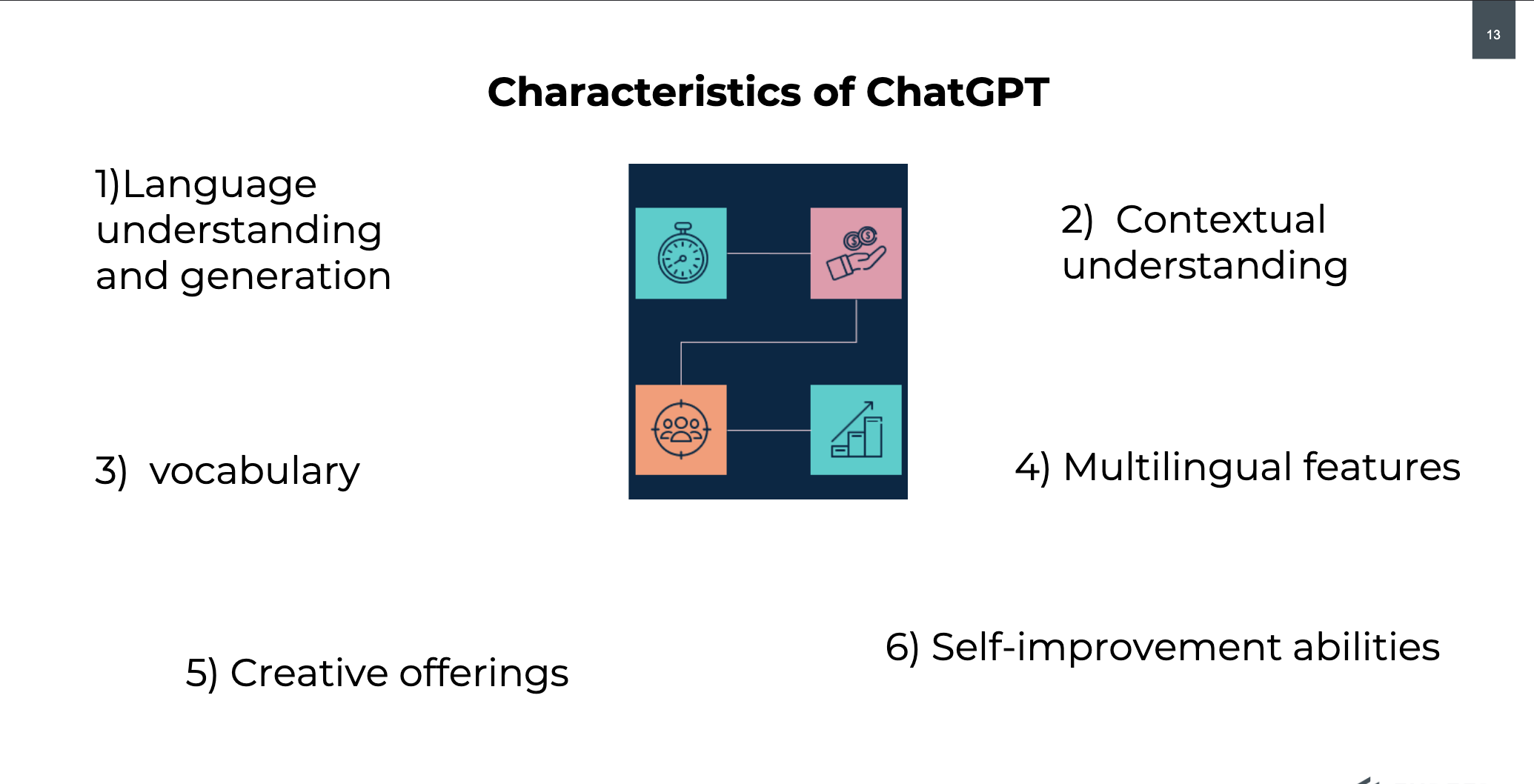}
  \caption{A PowerPoint presentation on education in terminology}
  \label{fig:Presentation_PowerPoint}
\end{figure}

{ % 开始一个局部作用域
\small % 设置字体大小
\begin{lstlisting}[caption=Example LDL for Characteristics Slide, basicstyle=\ttfamily\footnotesize, breaklines=true, escapechar=|]
<SOS>
SLIDE_CONTENT_TWO_COL ATTR_CENTER_IMAGE <SEP> /* Slide type indicating two columns with a central image/diagram */

/* Header Section */
ELEM_TITLE ATTR_TEXT_LENGTH_MEDIUM POS_TOP POS_CENTER <SEP> /* Characteristics of ChatGPT */

/* Central Diagram/Image */
ELEM_IMAGE ATTR_IMAGE_ASPECT_TALL ATTR_SIZE_PRIMARY POS_CENTER_HORIZONTAL POS_CENTER_VERTICAL <SEP> /* The central diagram */

/* Left Column Text Items */
ELEM_TEXT_BODY ATTR_TEXT_POINTS_FEW ATTR_TEXT_LENGTH_MEDIUM POS_MIDDLE_LEFT_UPPER <SEP> /* 1) Language understanding and generation */
ELEM_TEXT_BODY ATTR_TEXT_POINTS_FEW ATTR_TEXT_LENGTH_SHORT POS_MIDDLE_LEFT_CENTER <SEP> /* 3) vocabulary */
ELEM_TEXT_BODY ATTR_TEXT_POINTS_FEW ATTR_TEXT_LENGTH_MEDIUM POS_MIDDLE_LEFT_LOWER <SEP> /* 5) Creative offerings */

/* Right Column Text Items */
ELEM_TEXT_BODY ATTR_TEXT_POINTS_FEW ATTR_TEXT_LENGTH_MEDIUM POS_MIDDLE_RIGHT_UPPER <SEP> /* 2) Contextual understanding */
ELEM_TEXT_BODY ATTR_TEXT_POINTS_FEW ATTR_TEXT_LENGTH_MEDIUM POS_MIDDLE_RIGHT_CENTER <SEP> /* 4) Multilingual features */
ELEM_TEXT_BODY ATTR_TEXT_POINTS_FEW ATTR_TEXT_LENGTH_MEDIUM POS_MIDDLE_RIGHT_LOWER <SEP> /* 6) Self-improvement abilities */

/* Optional: Footer/Page Number if consistently present and part of design */
/* ELEM_FOOTER ATTR_TEXT_LENGTH_SHORT POS_BOTTOM_RIGHT <SEP> */ /* For page number, if treated as a design element */

<EOS>
\end{lstlisting}
} % 结束局部作用域，字体大小恢复

\subsection{Design Philosophy and Scope}
\label{appendix:ldl_philosophy}
LDL is intentionally designed to be a high-level, symbolic language. It abstracts away from precise coordinates and focuses on:
\begin{enumerate}
    \item \textbf{Semantic Content Types}: Distinguishing between titles, text, images, etc. 
    \item \textbf{Content Characteristics}: Capturing attributes like text length or image aspect ratio that influence layout choices .
    \item \textbf{General Zones and Sizing}: Specifying approximate locations and relative importance using position and size attributes.
\end{enumerate}

This approach contrasts with languages that define exact pixel positions or complex inter-element relational constraints (e.g., "element A is 10px to the left of element B"). While such precision is necessary for final rendering, LDL defers these fine-grained decisions to the Iterative Multi-Modal Optimization loop. In this loop, an initial Structured Intermediate Representation (SIR) is derived from the LDL, and then critics identify misalignments or aesthetic issues, which the Refinement Agent addresses using parameterized editing primitives.

\subsection{Limitations and Future Directions}
\label{appendix:ldl_limitations}
The current LDL vocabulary is expressive for common presentation layouts. However, highly complex or unconventional layouts (e.g., intricate infographics, non-grid-based designs) might require extensions to the vocabulary. Potential extensions could include more sophisticated grouping tokens (e.g., to define a set of elements that should be treated as a single visual block) or more explicit relative positioning tokens (e.g., \texttt{POS\_BELOW\_PREVIOUS}, \texttt{ALIGN\_WITH\_ELEMENT\_X}).

Future work could explore learning these extensions automatically from data or incorporating a richer set of relational primitives if deemed necessary for specific advanced use cases. Such advancements would need to be balanced against the increased complexity they might introduce for the LPG's learning task. The current design prioritizes learnability for the LPG and provides a robust symbolic starting point for the powerful iterative refinement process of the IMR loop.

\section{Layout Prototype Generator (LPG) Implementation Details}
\label{appendix:lpg_implementation_revised}

This appendix provides key implementation specifications for the Layout Prototype Generator (LPG). The LPG transforms slide concept features into symbolic Layout Description Language (LDL) sequences using a Transformer encoder-decoder architecture.

\subsection{Input Feature Representation}
\label{appendix:lpg_input_features_revised}
The input to the LPG encoder for each slide concept is a 512-dimensional vector, derived from the concatenation and projection of content-derived features and categorical metadata features.
\begin{itemize}
    \item \textbf{Content Features ($feat_{content,j}$):} Textual content ($T_{ij}$, bullet points from R-CoT) is encoded using a RoBERTa-base model, and associated images ($F_{ij}$) are encoded using a ViT-B/16 model. These features are individually projected and then concatenated to form a combined content feature vector (256-dimensions). If no image is present, a zero vector is used for the visual component.
    \item \textbf{Categorical Features ($type_j$ and other metadata):} These include the slide's functional type (10 categories), estimated number of bullet points (3 categories: `few`, `medium`, `many`), and primary image aspect ratio (3 categories: `wide`, `square`, `tall`). Each feature is mapped to an integer index and passed through separate embedding layers. The resulting embeddings are concatenated to form a categorical feature vector (64-dimensions).
    \item \textbf{Final Input Embedding:} The 256-dim content feature vector and the 64-dim categorical feature vector are concatenated (320-dim total) and then projected to the Transformer's model dimension of 512 via a final linear layer.
\end{itemize}

\subsection{Transformer Architecture Details}
\label{appendix:lpg_architecture_revised}
LPG employs a standard Transformer encoder-decoder architecture.
\begin{itemize}
    \item \textbf{Shared Parameters:} Both encoder and decoder utilize 6 layers, 8 attention heads, a model hidden dimension ($d_{\text{model}}$) of 512, and a feed-forward network (FFN) inner dimension of 2048 with GELU activation. Layer Normalization (Pre-LN) and a dropout rate of 0.1 are applied consistently.
    \item \textbf{Encoder Specifics:} Standard sinusoidal positional encodings are added to the input embeddings (input treated as a sequence of length 1).
    \item \textbf{Decoder Specifics:} Standard sinusoidal positional encodings are added to the target LDL token embeddings. The output layer consists of a linear projection to the LDL vocabulary size (200 tokens), followed by a Softmax function.
\end{itemize}
\noindent \textbf{Total Trainable Parameters:} Approximately 44.5 Million.

\subsection{Training Procedure}
\label{appendix:lpg_training_revised}
The LPG is trained as follows:
\begin{itemize}
    \item \textbf{Objective Function:} Cross-entropy imitation loss with L2 weight decay (coefficient $\alpha_{\text{L2}}=10^{-4}$).
    \item \textbf{Optimizer \& Learning Rate:} AdamW optimizer with a peak learning rate of $3 \times 10^{-4}$, using a linear warm-up for the first 10\% of training steps followed by a cosine annealing decay to $1 \times 10^{-5}$.
    \item \textbf{Batching \& Regularization:} Batch size of 64; gradient clipping with a maximum L2 norm of 1.0.
    \item \textbf{Data \& Duration:} Trained on the Zenodo10K subset (see Section 5.1 of the main paper) for up to 50 epochs, with early stopping (patience of 5 epochs) based on validation loss.
    \item \textbf{Infrastructure:} 4 NVIDIA A100 (40GB) GPUs, using PyTorch and the Hugging Face Transformers library.
\end{itemize}

\subsection{Inference (LDL Generation)}
\label{appendix:lpg_inference_revised}
For generating LDL sequences at inference time:
\begin{itemize}
    \item \textbf{Decoding Strategy:} Beam Search.
    \item \textbf{Beam Size:} 5.
    \item \textbf{Maximum Sequence Length:} 128 tokens.
\end{itemize}

\section{Editing Primitives}
\label{appendix:editing_primitives}

This appendix details the Editing Primitives (EPs) used in the Iterative Multi-Modal Optimization process of RCPS framework. These primitives provide a controlled interface for the Refinement Agent to modify presentations based on critique feedback.

\subsection{Introduction}
Editing Primitives are deterministic functions used by the Refinement Agent (LLM) to modify a structured intermediate representation (SIR) of the slide, based on critiques from VLM-C and LLM-C. The SIR contains objects for each slide element with precise geometric, style, and content attributes. EPs provide a controlled mechanism for iterative refinement.

\subsection{Positional and Alignment Primitives}

\paragraph{move\_element(id, dx, dy)}
\begin{description}
    \item[Purpose:] Translates the element specified by \texttt{id}.
    \item[Parameters:]
    \begin{itemize}
        \item \texttt{id} (string): Unique identifier of the target element.
        \item \texttt{dx} (float): Horizontal displacement (unit: pixels).
        \item \texttt{dy} (float): Vertical displacement (unit: pixels).
    \end{itemize}
    \item[Effect:] Updates the \texttt{x}, \texttt{y} coordinates of the element in the SIR.
\end{description}

\paragraph{adjust\_alignment(id, reference\_id, alignment\_type)}
\begin{description}
    \item[Purpose:] Aligns the target element relative to a reference object.
    \item[Parameters:]
    \begin{itemize}
        \item \texttt{id} (string): Unique identifier of the element to align.
        \item \texttt{reference\_id} (string): Identifier of the reference element, or special values \texttt{"slide\_bounds"}, \texttt{"slide\_center"}.
        \item \texttt{alignment\_type} (string): Specifies the alignment type. Implemented values: \texttt{'left'}, \texttt{'right'}, \texttt{'top'}, \texttt{'bottom'}, \texttt{'center\_h'}, \texttt{'center\_v'}.
    \end{itemize}
    \item[Effect:] Calculates required displacement and calls \texttt{move\_element} to update the element's position in the SIR.
\end{description}

\subsection{Sizing Primitives}

\paragraph{resize\_element(id, dw, dh, anchor\_point='center')}
\begin{description}
    \item[Purpose:] Changes the width and height of the specified element.
    \item[Parameters:]
    \begin{itemize}
        \item \texttt{id} (string): Unique identifier of the target element.
        \item \texttt{dw} (float): Change in width (unit: pixels).
        \item \texttt{dh} (float): Change in height (unit: pixels).
        \item \texttt{anchor\_point} (string, default=\texttt{'center'}): The fixed point during resizing. Values include: \texttt{'center'}, \texttt{'top\_left'}, \texttt{'top\_right'}, \texttt{'bottom\_left'}, \texttt{'bottom\_right'}, \texttt{'middle\_left'}, \texttt{'middle\_right'}, \texttt{'top\_center'}, \texttt{'bottom\_center'}.
    \end{itemize}
    \item[Effect:] Updates the \texttt{w}, \texttt{h} attributes (and possibly \texttt{x}, \texttt{y} depending on \texttt{anchor\_point}) of the element in the SIR.
\end{description}

\subsection{Content Modification Primitives (Text)}

\paragraph{rewrite\_bullet\_point(id, index, new\_text)}
\begin{description}
    \item[Purpose:] Replaces the text of a specific part within a text element.
    \item[Parameters:]
    \begin{itemize}
        \item \texttt{id} (string): Unique identifier of the text element.
        \item \texttt{index} (integer): Zero-based index of the text part to modify.
        \item \texttt{new\_text} (string): The new text content.
    \end{itemize}
    \item[Effect:] Modifies the internal text storage of the element in the SIR.
\end{description}

\paragraph{delete\_bullet\_point(id, index)}
\begin{description}
    \item[Purpose:] Removes a specific part of the text from a text element.
    \item[Parameters:]
    \begin{itemize}
        \item \texttt{id} (string): Unique identifier of the text element.
        \item \texttt{index} (integer): Zero-based index of the part to delete.
    \end{itemize}
    \item[Effect:] Removes the specified content from the element's text storage in the SIR.
\end{description}

\subsection{Style and Formatting Primitives}

\paragraph{change\_style(id, attribute, value)}
\begin{description}
    \item[Purpose:] Modifies a single visual style attribute of an element.
    \item[Parameters:]
    \begin{itemize}
        \item \texttt{id} (string): Unique identifier of the target element.
        \item \texttt{attribute} (string): Name of the style attribute. Supported attributes in this implementation: \texttt{'font\_size'}, \texttt{'font\_weight'}, \texttt{'font\_color'}, \texttt{'fill\_color'}, \texttt{'border\_color'}, \texttt{'border\_width'}, \texttt{'text\_alignment'}, \texttt{'opacity'}.
        \item \texttt{value} (any): The new value for the attribute.
    \end{itemize}
    \item[Effect:] Updates the specified style attribute value for the element in the SIR.
\end{description}

\paragraph{recolor\_element(id, property, color\_value)}
\begin{description}
    \item[Purpose:] Specifically modifies color attributes.
    \item[Parameters:]
    \begin{itemize}
        \item \texttt{id} (string): Unique identifier of the target element.
        \item \texttt{property} (string): Specifies color aspect: \texttt{'fill'}, \texttt{'text'}, \texttt{'border'}.
        \item \texttt{color\_value} (string): The new color value (format: \texttt{'\#RRGGBB'}).
    \end{itemize}
    \item[Effect:] Updates the corresponding color attribute in the SIR.
\end{description}

\paragraph{reformat\_text(id, style\_params)}
\begin{description}
    \item[Purpose:] Applies multiple text formatting changes simultaneously.
    \item[Parameters:]
    \begin{itemize}
        \item \texttt{id} (string): Unique identifier of the text element.
        \item \texttt{style\_params} (dict): Dictionary of style attributes and their new values.
    \end{itemize}
    \item[Effect:] Internally calls \texttt{change\_style} for each item in \texttt{style\_params}.
\end{description}

\subsection{Spacing Primitive}

\paragraph{adjust\_spacing(id1, id2, target\_space, direction)}
\begin{description}
    \item[Purpose:] Sets the spacing between the bounding boxes of two specified elements.
    \item[Parameters:]
    \begin{itemize}
        \item \texttt{id1}, \texttt{id2} (string): Identifiers of the two elements.
        \item \texttt{target\_space} (float): Desired space between elements (unit: pixels).
        \item \texttt{direction} (string): \texttt{'horizontal'} or \texttt{'vertical'}.
    \end{itemize}
    \item[Effect:] Calculates current spacing, determines required displacement, and calls \texttt{move\_element} to update the SIR.
\end{description}

\section{Critique Format Example}
\label{appendix:critique_format}
An example of the structured JSON feedback format used by the Visual Critic:
\begin{lstlisting}
{
  "issues": [
    {
      "element_id": "title",
      "issue_type": "Misalignment",
      "severity": 0.75,
      "target_element_id": "slide_bounds",
      "suggestion": "Center the title horizontally"
    },
    {
      "element_id": "bullet_list",
      "issue_type": "Overflow",
      "severity": 0.9,
      "suggestion": "Reduce font size or content length"
    }
  ]
}
\end{lstlisting}

\section{PREVAL Evaluation Workflow Algorithm}
\label{appendix:pefval_workflow_algo}

This appendix details the PREVAL evaluation workflow algorithm.

\begin{algorithm}[H] 
  \caption{PREVAL Evaluation Workflow}
  \label{alg:preval} 
  \begin{algorithmic}[1]
    \REQUIRE Presentation PPT to evaluate, Pre-trained assessment functions $\{q_k^*\}_{k \in \mathcal{K}}$
    \ENSURE Dimensional quality scores $\{\text{Score}_k\}_{k \in \mathcal{K}}$, Aggregate score $\text{Score}_{\text{PREVAL}}$
    \STATE \textbf{Step 1:} Extract multi-modal feature representation
      \STATE $x_{\text{PPT}} \gets \phi_{\text{model}}(\text{parse}(\text{PPT}), \text{render}(\text{PPT}))$
    \STATE \textbf{Step 2:} Score along each quality dimension
      \FOR{each dimension $k \in \mathcal{K}$}
        \STATE $\text{raw\_score}_k \gets q_k^*(x_{\text{PPT}})$
        \STATE $\text{Score}_k \gets \text{Normalize}(\text{raw\_score}_k)$
      \ENDFOR
    \STATE \textbf{Step 3:} Calculate aggregate assessment
      \STATE $\text{Score}_{\text{PREVAL}} \gets \sum_{k \in \mathcal{K}} w_k \cdot \text{Score}_k$
    \RETURN $\{\text{Score}_k\}_{k \in \mathcal{K}}$, $\text{Score}_{\text{PREVAL}}$
  \end{algorithmic}
\end{algorithm}

\section{Human Evaluation Protocol}
\label{app:human_eval_protocol_revised_en}

This appendix details the protocol followed for all human evaluation tasks conducted in this study, including the annotation of the PREVAL Preference Dataset (Section H.2) and the direct human evaluation of generated presentations (Section H.3 and Section 5). The goal was to establish a rigorous and consistent methodology for assessing presentation quality.

\subsection{Personnel Recruitment and Training}

\begin{itemize}
    \item \textbf{Recruitment Criteria:} We recruited five professionals. All professionals were required to possess:
        \begin{enumerate}
            \item A Master's degree.
            \item Practical work experience in academic or business fields involving the creation or frequent use of presentations.
            \item Demonstrable experience in creating presentations using standard software (Microsoft PowerPoint, Google Slides).
        \end{enumerate}
    \item \textbf{Training and Calibration:}
        \begin{enumerate}
            \item \textbf{Project Briefing (1 hour):} Professionals were provided with an overview of the project, the objectives of automated presentation generation, and the significance of their role in quality assessment. Key concepts such as "Content Relevance," "Logical Coherence," and "Visual Design" were introduced with illustrative examples of effective and ineffective practices.
            \item \textbf{Guideline Study and Q\&A (Self-paced + 1-hour Q\&A):} Detailed evaluation guidelines (summarized below) were distributed. Professionals studied these guidelines independently, followed by a 1-hour question and answer session with the researchers to resolve any queries.
            \item \textbf{Calibration Session (2 hours):} A set of 12 sample presentation pairs (for preference tasks) and 5 full presentations (for Likert scale rating), not part of the main study data, were used for calibration.
            Professionals first independently completed evaluations for these samples.
            Subsequently, their evaluations were discussed collectively with the research team. Significant discrepancies in ratings or rationales were analyzed, and a consensus was established regarding the evaluation criteria and consistent application of rating scales/preference judgments.
            Particular emphasis was placed on differentiating between the three quality dimensions (Content, Coherence, Design) to minimize assessment biases.
            \item \textbf{Pilot Task:} Before commencing the main evaluation, professionals completed a pilot task involving 20 preference pairs and 2 full presentations, and received specific feedback.
        \end{enumerate}
\end{itemize}

\subsection{Evaluation Task 1: PREVAL Preference Dataset Annotation (Pairwise Comparisons)}
\label{app:Dataset Annotation}

\begin{itemize}
    \item \textbf{Task Objective:} For each pair of presentations $(\text{PPT}_A, \text{PPT}_B)$ derived from the same source document, professionals provided preference judgments and rationales across three dimensions.
    \item \textbf{Interface:} A custom web-based annotation interface was used, displaying $\text{PPT}_A$ and $\text{PPT}_B$ side-by-side, with access to the source document.
    \item \textbf{Dimensions and Judgments:} For each dimension ($k \in \{\text{Content, Coherence, Design}\}$):
        \begin{itemize}
            \item \textbf{Preference:} Select one of the following five levels:
            `A is Significantly Better than B` | `A is Slightly Better than B` | `B is Significantly Better than A` | `B is Slightly Better than A` | `A and B are of Similar Quality`
            \item \textbf{Rationale (Mandatory):}
                \begin{itemize}
                    \item Provide 1-3 sentences of free-text explanation justifying the preference.
                    \item Select up to 3 predefined structured tags from a provided list that best describe the reasons for the preference (e.g., for Coherence: "Clearer transitions in A", "B lacks logical flow"). The tag list was iteratively developed based on an analysis of common issues in automatically generated presentations.
                \end{itemize}
        \end{itemize}
    \item \textbf{Guidance for Dimensions:}
        \begin{itemize}
            \item \textbf{Content:} Focus on the accuracy and completeness of key information from the source, relevance of content to slide themes, and avoidance of information redundancy or fabrication.
            \item \textbf{Coherence:} Evaluate the logical flow between slides, clarity of transitions, overall narrative structure, and whether the presentation forms a cohesive whole.
            \item \textbf{Design:} Assess visual appeal, professionalism, layout appropriateness for the content, readability (fonts, text size, contrast), use of white space, alignment, consistency in visual style, and image quality/relevance.
        \end{itemize}
    \item \textbf{Annotation Process:} Each presentation pair was evaluated by three different professionals.
\end{itemize}

\subsection{Evaluation Task 2: Direct Human Evaluation of Full Presentations (Likert Scale Rating)}

\begin{itemize}
    \item \textbf{Task Objective:} To obtain absolute quality ratings for full presentations generated by RCPS and baseline methods.
    \item \textbf{Interface:} Professionals viewed each full presentation sequentially (PDF or slide show format), with access to the source document.
    \item \textbf{Evaluation Aspects and Scale:} Professionals rated each presentation on a 7-point Likert scale (1=Very Poor, 4=Average, 7=Excellent) for the following four aspects:
        \begin{enumerate}
            \item \textbf{Content Relevance \& Accuracy (Mapped to PREVAL-Content):}
                \begin{itemize}
                    \item \textit{1 (Very Poor):} Content is irrelevant, inaccurate, or omits most key information.
                    \item \textit{4 (Average):} Content relevance and accuracy are average; captures some key information but has omissions or inaccuracies.
                    \item \textit{7 (Excellent):} Content is highly relevant, accurate, fully captures key information from the source, and is well-summarized.
                \end{itemize}
            \item \textbf{Logical Flow \& Coherence (Mapped to PREVAL-Coherence):}
                \begin{itemize}
                    \item \textit{1 (Very Poor):} Presentation is difficult to understand, lacks logic, slides are disjointed.
                    \item \textit{4 (Average):} Narrative flow is generally understandable, but transitions may be unnatural or connections unclear.
                    \item \textit{7 (Excellent):} Presentation has a clear, logical, and smooth narrative flow.
                \end{itemize}
            \item \textbf{Visual Appropriateness \& Design (Mapped to PREVAL-Design):}
                \begin{itemize}
                    \item \textit{1 (Very Poor):} Design is unprofessional, visually poor, layout is inappropriate, text is illegible.
                    \item \textit{4 (Average):} Design is acceptable but unexceptional; layout is functional but may have aesthetic flaws.
                    \item \textit{7 (Excellent):} Design is highly professional, visually appealing, layout is excellent and aids content understanding.
                \end{itemize}
            \item \textbf{Overall Satisfaction:}
                \begin{itemize}
                    \item \textit{1 (Very Poor):} Very dissatisfied; presentation is ineffective and of low quality.
                    \item \textit{4 (Average):} Neither satisfied nor dissatisfied; presentation is mediocre.
                    \item \textit{7 (Excellent):} Very satisfied; presentation is effective, engaging, and of high quality.
                \end{itemize}
        \end{enumerate}
    \item \textbf{Evaluation Process:} Each presentation was independently evaluated by all five professionals. Professionals were encouraged to add brief optional comments for outlier scores or specific strengths/weaknesses.
\end{itemize}

\subsection{Ensuring Evaluation Quality}

\begin{itemize}
    \item \textbf{Regular Communication:} Weekly brief meetings were held with professionals during the main evaluation phase to address queries and maintain consistency.
    \item \textbf{Quality Checks:} Researchers periodically reviewed a small percentage of evaluations to monitor quality and provide feedback if necessary.
    \item \textbf{Inter-Rater Reliability (IRR):} As reported in Section H.2 and H.3, IRR was calculated (Fleiss' Kappa for preference judgments, Krippendorff's Alpha for Likert scale ratings) to confirm the reliability of the collected human judgments. Evaluation items with low initial agreement were subject to review and discussion.
\end{itemize}

This protocol was designed to maximize the consistency, reliability, and validity of the human judgments collected for this research.

\section{Additional Result Visualizations}
\label{app:main_result_viz}
% Content for this appendix section

\begin{figure}[h]
  \centering
  \includegraphics[width=0.32\columnwidth]{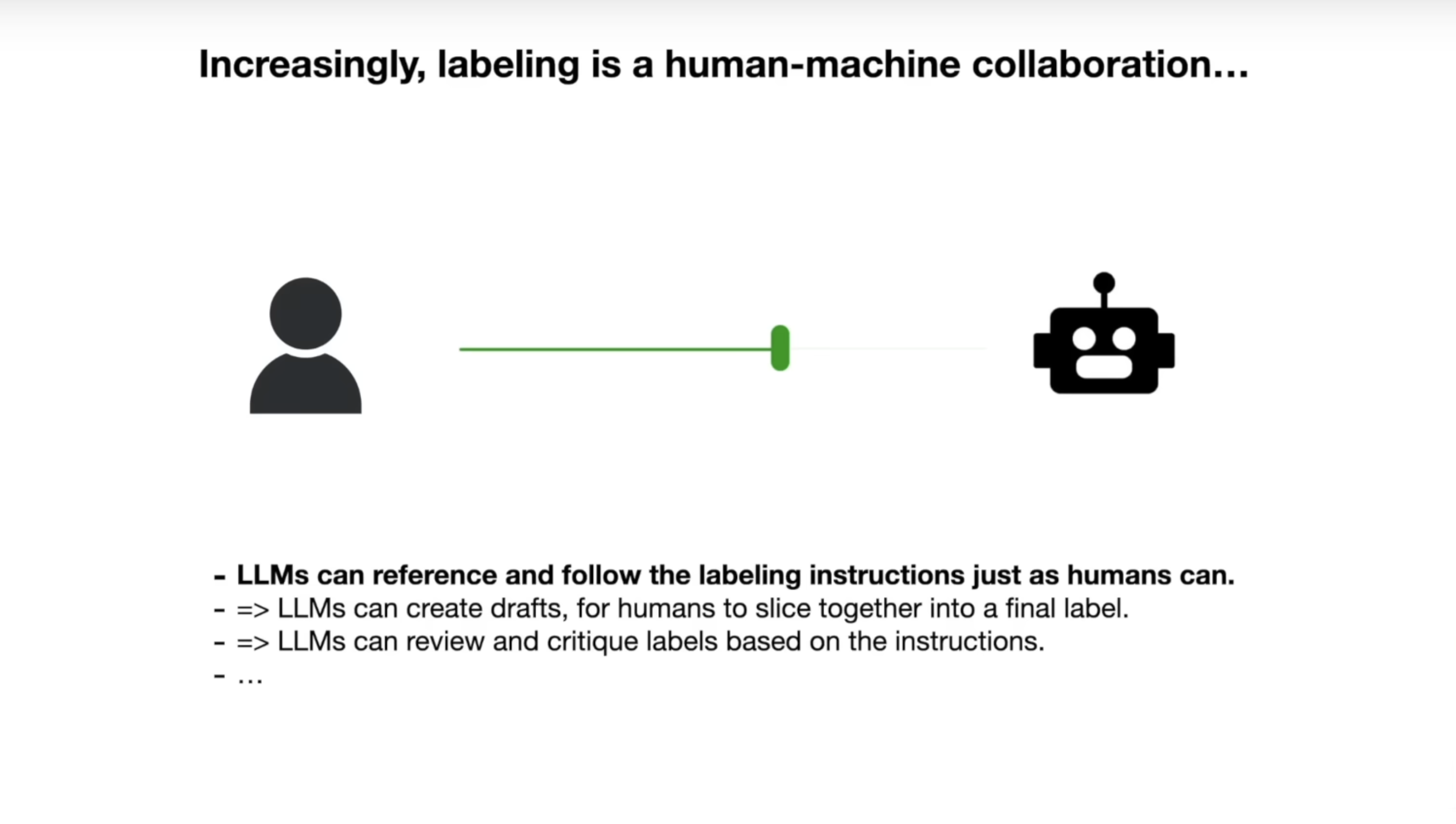}
  \hfill
  \includegraphics[width=0.32\columnwidth]{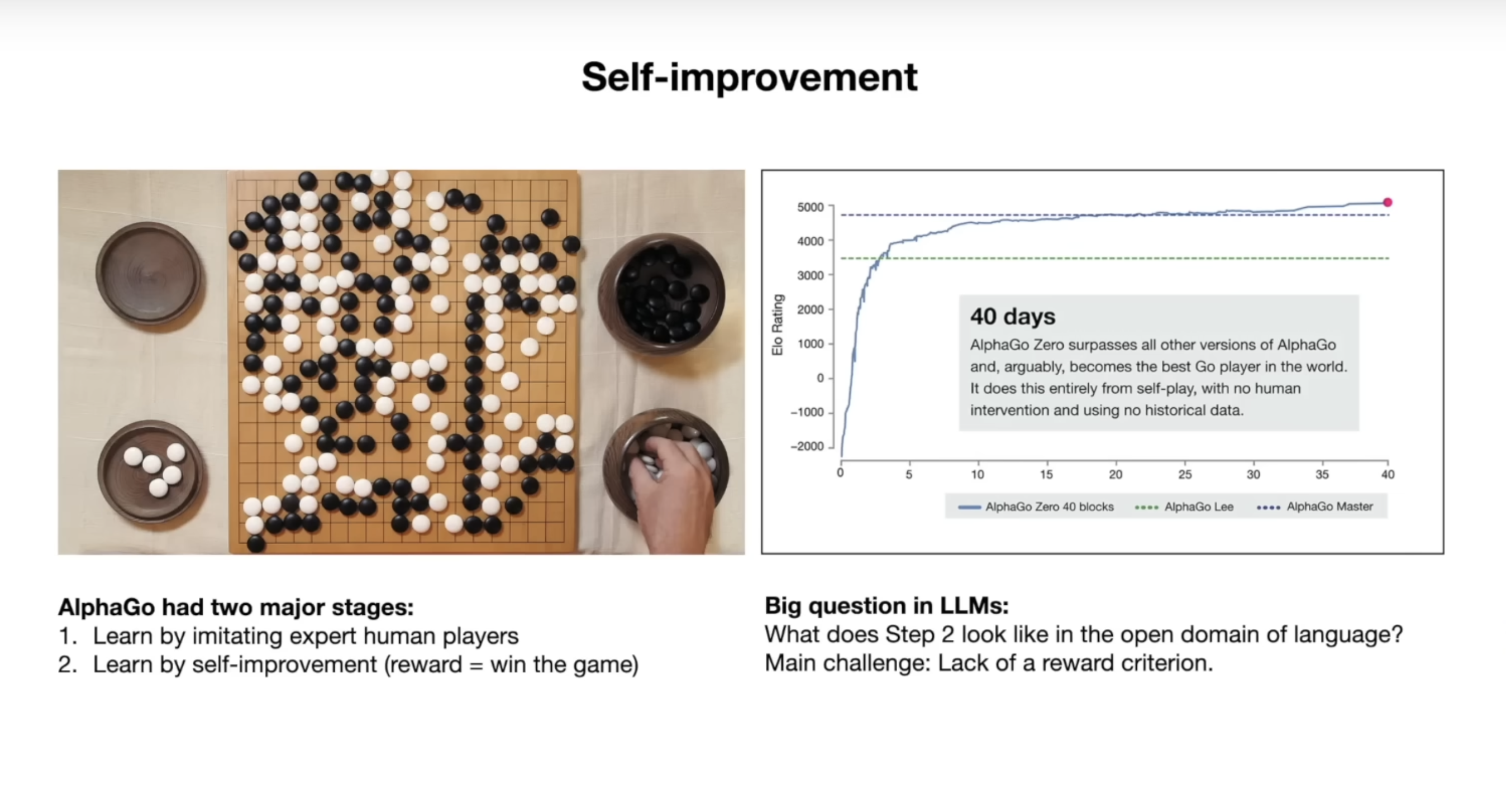}
  \hfill
  \includegraphics[width=0.32\columnwidth]{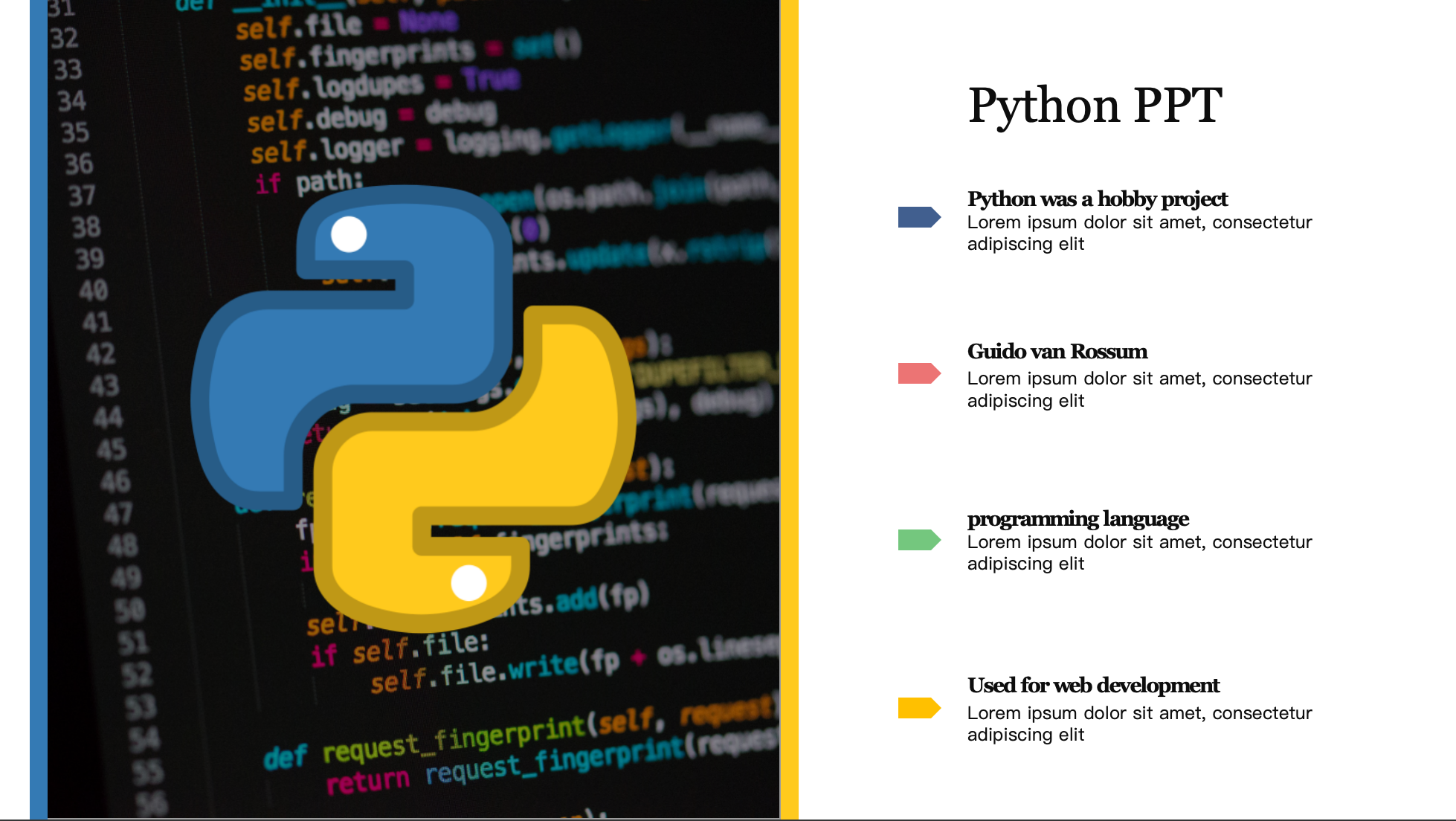}
  \caption{Some additional result visualizations}
  \label{fig:multiple_results}
\end{figure}

\end{document}